\documentclass[10pt,twocolumn,letterpaper]{article}
\usepackage[pagenumbers]{paper}
\usepackage{xcolor}
\usepackage{multirow}
\usepackage{makecell}
\usepackage{threeparttable}
\usepackage{booktabs}
\usepackage[pagebackref,breaklinks,colorlinks,allcolors=linkblue]{hyperref}
\usepackage[normalem]{ulem}
\usepackage{marvosym}
\usepackage{xspace}

\usepackage{xurl}

\definecolor{linkblue}{rgb}{0.21,0.49,0.74}

\newcommand{\ours}{\textit{PathoSpatial}\xspace}

\title{Prototype-driven fusion of pathology and spatial transcriptomics for interpretable survival prediction}

\author{
    \begin{tabular}{c}
        Lihe Liu \textsuperscript{1,3},
        Xiaoxi Pan \textsuperscript{2,3}, 
        Yinyin Yuan \textsuperscript{2,3,$\dagger$ \Letter}, 
        Lulu Shang \textsuperscript{1,3,$\dagger$ \Letter}\\[0.5em]
        \normalsize \textsuperscript{1} Department of Biostatistics, MD Anderson Cancer Center, Houston, TX \\
        \normalsize \textsuperscript{2} Department of Translational Molecular Pathology, MD Anderson Cancer Center, Houston, TX \\
        \normalsize \textsuperscript{3} The Institute for Data Science in Oncology (IDSO), MD Anderson Cancer Center, Houston, TX \\
        {\tt\small \{yyuan6, lshang\}@mdanderson.org }
    \end{tabular}
}

\begin{document}
\maketitle
\let\thefootnote\relax\footnotetext{\textsuperscript{$\dagger$} Corresponding authors. Equal supervision.}

\begin{abstract}

Whole slide images (WSIs) enable weakly supervised prognostic modeling via multiple instance learning (MIL). Spatial transcriptomics (ST) preserves in situ gene expression, providing a spatial molecular context that complements morphology. As paired WSI-ST cohorts scale to population level, leveraging their complementary spatial signals for prognosis becomes crucial; however, principled cross-modal fusion strategies remain limited for this paradigm. To this end, we introduce $\ours$, an interpretable end-to-end framework integrating co-registered WSIs and ST to learn spatially informed prognostic representations. $\ours$ uses task-guided prototype learning within a multi-level experts architecture, adaptively orchestrating unsupervised within-modality discovery with supervised cross-modal aggregation. By design, $\ours$ substantially strengthens interpretability while maintaining discriminative ability. We evaluate $\ours$ on a triple-negative breast cancer cohort with paired ST and WSIs. $\ours$ delivers strong and consistent performance across five survival endpoints, achieving superior or comparable performance to leading unimodal and multimodal methods. $\ours$ inherently enables post-hoc prototype interpretation and molecular risk decomposition, providing quantitative, biologically grounded explanations, highlighting candidate prognostic factors. We present $\ours$ as a proof-of-concept for scalable and interpretable multimodal learning for spatial omics–pathology fusion.

\end{abstract}
    
\section{Introduction}
\label{sec:intro}

\begin{figure}[t]
  \centering
  \includegraphics[width=1\linewidth]{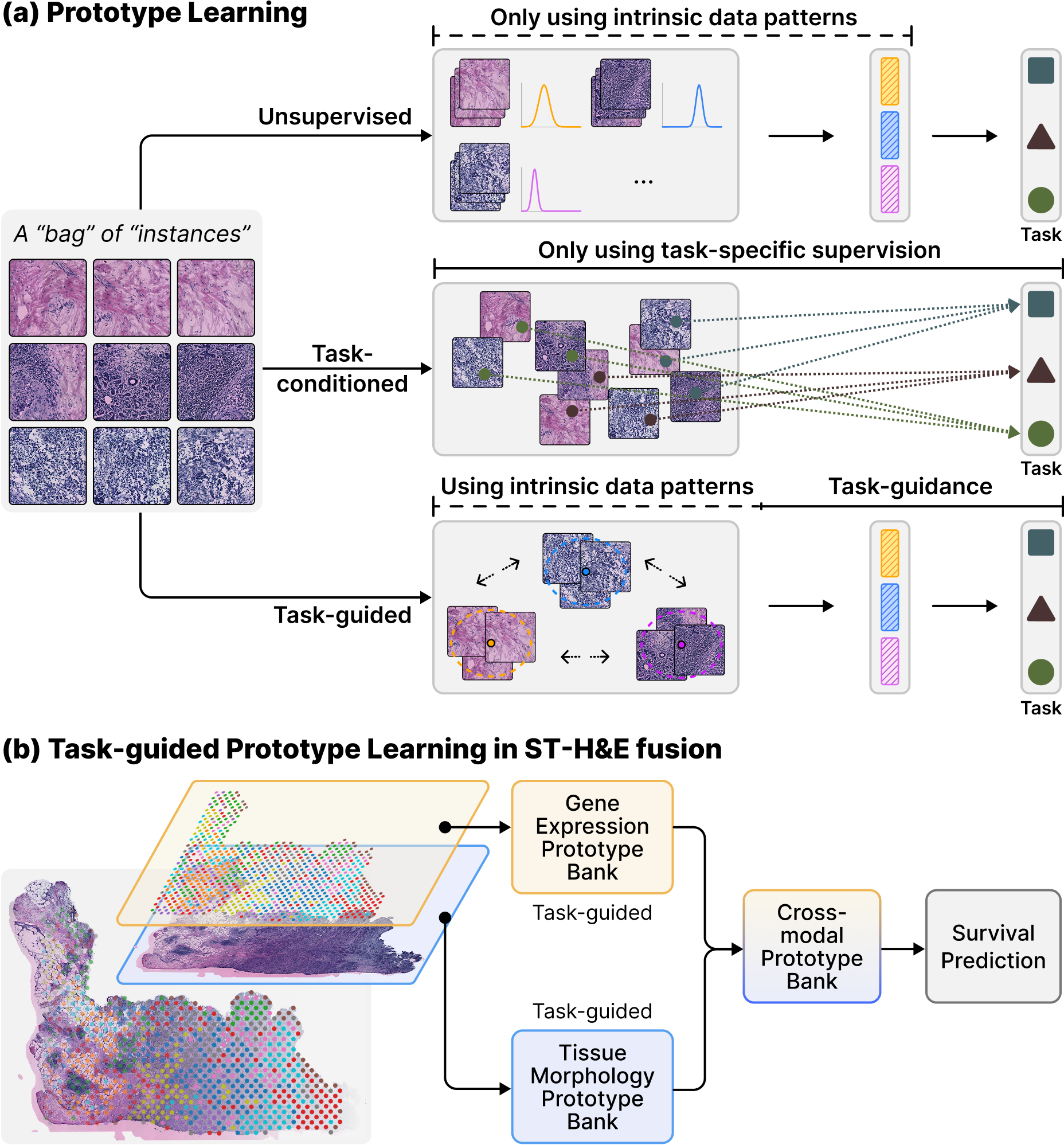}

   \caption{
    \textbf{Task-guided Prototype Learning}. (a) Prototype-based methods provide compact and effective representations of WSIs in MIL. Prior strategies have generally taken two forms, focusing on interpretability (fully unsupervised) or task-alignment (fully task-conditioned). We explore a hybrid task-guided prototype learning strategy balancing interpretability and task-guidance. (b) We apply task-guided prototype learning in the fusion of ST and H\&E.
    }
   \label{fig:fig1_task_guided_proto}

\end{figure}
Whole Slide Images (WSIs) provide critical histological insights that are indicative of disease progression and patient outcome, which has enabled powerful computer-aided analysis in the field of computational pathology (CPath) \cite{li2022cpathoverview, campanella2019clinical, wulczyn2021interpretable, yu2016nsclsprognosis}. Conventionally, the challenges posed by the gigapixel resolution and the lack of pixel-level annotations are addressed using Multiple Instance Learning (MIL) \cite{abmil2018, campanella2019clinical, chen2013mil, amores2013mil}, which models a WSI as a ``bag" of ``instances" to enable slide-level inference in a weakly supervised learning approach. Emerging multi-modal methods that integrate WSIs with genomics data have shown enhanced accuracy of patient outcome predictions \cite{shao2019multimodel1, chen2021porpoise, jaume2023survpath, song2024multimodal}, however, a primary limitation is the reliance on bulk sequencing data which pools genomic signals from heterogeneous cells, masking localized gene expression and spatial patterns vital for identifying critical cellular niches in the tumor microenvironment (TME). 

Spatial Transcriptomic (ST) technology has been made more accessible \cite{stahl2016st1ST, he2022st2Cosmx, merritt2020st3Geomx}, which is designed to measure spatially resolved gene expressions from the intact tissue section. Measuring gene expression with preserved physical context enables direct quantification of spatial context on cellular phenotype and tissue function, such spatially resolved view has proved especially informative in prognostication. For example, identification of core versus leading-edge expression signatures associated with survival and targeted therapy response \cite{arora2023spatialtumoredge}, spatial archetypes associated with patient survival \cite{wang2024spatialtnbc}. Collectively, these finer markers link patient outcomes to both molecular states and their spatial distribution. Thus, integrating ST with H\&E in prognostic models offers a promising path to improved accuracy and interpretability, yet effective fusion strategies for these two modalities remain underexplored.

As ST studies transition from modest sample sizes into an era of large-scale cohort initiatives \cite{cooper2025bone_xenium, grouard2025spamil, wall2025smart}, ongoing advances in hardware throughput and declining costs will surely continue to accelerate this trajectory, marking a pivotal shift in the data landscape. This transition underscores the critical and timely need for specialized computational algorithms to scale with these expanding datasets. Leveraging ST for weakly supervised prognostic modeling presents unique constraints. ST is fine-grained yet spatially sparse, leading to substantial high-resolution noise and redundancy that must be separated from truly discriminative signals, e.g., prototype learning. Moreover, pairing ST with heterogeneous morphology introduces complex, sample-dependent cross-modal interactions, where the correspondence is often nonlinear and dynamic. Static fusion that assumes a fixed global alignment may therefore amplify modality-specific noise or dilute complementary signals. This motivates a framework that effectively learns highly discriminative representations while performing adaptive fusion beyond static fusion.

Prototype learning has been increasingly used in ``panoramic" CPath MIL tasks to reduce redundancy and aggregate vast patches into a small set of contributing ``concepts" prototypes \cite{song2024panther, zhang2024pibd, LiuFen2025prosurv}, which enables effective information compression, reducing training burden \cite{song2024panther, song2024multimodal} and facilitating cross-modal alignment \cite{zhang2024pibd, LiuFen2025prosurv}. Conventional prototype learning typically uses either unsupervised aggregation or task-conditioned aggregation (Fig. \ref{fig:fig1_task_guided_proto}). Unsupervised methods excel at capturing comprehensive representations by learning intrinsic data structures; however, these task-agnostic features may absorb task-irrelevant/contradictory patterns and sample-specific noise. Fully task-conditioned methods optimize the task objective but may underrepresent tissue heterogeneity, blurring distinct patterns and complicating interpretation.

Based on these insights, we propose $\ours$, a multimodal learning framework to leverage population-level ST and paired WSI for patient survival predictions. We hypothesize that each modality consists of complementary information predictive of patient survival risk, the combination of tissue morphology signals and gene regulatory program can jointly improve the prediction performance. Inspired by the Mixture of Experts (MoE) architecture \cite{jacobs1991moe1, Shazeer2017moe2}, we develop a hierarchical multi-level experts framework consisting of (i) within-modality prototype experts, and (ii) a cross-modal fusion expert, sequentially. The within-modality experts identify modality specific prognostic patterns optimized by the task objective while preserving intrinsic data structures to enhance interpretability, the fusion expert adaptively integrates these cross modal prototypical signals to form a holistic and task aligned representation. This hierarchical architecture is designed for effective learning of discriminative features within each modality while enabling the adaptive fusion of complementary signals. To summarize, our contributions are as follows:

(1) The first principled end-to-end MIL framework tailored to paired ST–H\&E data. 

(2) Extensive experiments across state-of-the-art MIL backbones with different fusion operators on a large-scale cohort. Our method shows superior or competitive performance with adapted baselines, with notably high consistency across five endpoints, highlighting the strong potential despite currently limited external validation. 

(3) A novel post-hoc interpretability analysis enabled by our design, providing quantitative, biologically grounded explanations and highlighting candidate prognostic factors.
\section{Related Work}
\label{sec:related_work}

\subsection{MIL in Computational Pathology}
MIL aggregators can be categorized by their spatial information usage. \textbf{Spatial-unaware} methods treat a slide as a permutation-invariant set of patch embeddings and learn which instances matter without using position: (a) Key-instance selectors focus on a small subset of highly suspicious tiles and aggregate via max/top-k, simple and effective when signal is localized 
but fails to capture diffuse patterns \cite{campanella2019clinical}; (b) Attention pooling assigns soft weights to all patches to form a weighted bag representation \cite{abmil2018, lu2021clam, li2021dsmil} and (c) cluster-aware variants impose a phenotype structure before pooling (e.g., clustering high-score tiles, centroid reduction, latent-space augmentation) to reduce redundancy and stabilize learning in heterogeneous slides \cite{Yao_2020DeepAttnMISL, yao2019deepmisl}. \textbf{Spatial-aware} methods explicitly model slide layout and learn context-aware embeddings: (a) Self-attention models use all pairs interactions with positional encoding to capture long-range and multi-scale tissue organization \cite{shao2021transmil, zheng2023kat, zhao2022setmil}, and (b) Graph-based approaches construct tile-adjacency graphs and propagate local-to-global information, often coupled with attention or hierarchical pooling \cite{chen2021patchgcn, li2018deepgraphconv, zheng2022gtp, ding2023egt}.

\subsection{Multimodal Fusion in MIL} 

\textbf{Late fusion} approaches encode each modality independently and merge slide-level embeddings at the prediction stage, typically via concatenation or bilinear \cite{chen2020pathomic, chen2021porpoise}. Multimodal learning frameworks that combine histology with clinical variables and genomic information have shown a promising slide-level prognosis \cite{mobadersany2018predicting, chen2021porpoise, ding2023pathology, volinskyfremond2024prediction}. \textbf{Early-fusion} methods mix information before per-modality pooling by modeling token-level interactions across modalities, commonly using self-attention on joint token sets \cite{jaume2023survpath} or simplified cross-attention between different modalities 
\cite{chen2021mcat, Xu2023motcat, zhou2023CMTA}.

\subsection{Prototype-based Set Representation}

Emerging approaches have been proposed to better model tissue heterogeneity, primarily by learning compact descriptors from repeating histology patterns which reflect the same morphology. \textbf{Unsupervised approaches} derive task-agnostic bag representations by clustering (e.g., K-means \cite{vu2023h2t, Yao_2020DeepAttnMISL, hou2023hgt}, Gaussian mixture model \cite{song2024panther}, Leiden community detection \cite{claudioquiros2024hpl}) or graph models \cite{hou2023hgt, claudioquiros2024hpl, chan2023histopathology} without labels. The final slide representation often takes the form of compositional vectors, aggregated prototype features or pooled graph features, assuming that these structures capture internal biological variance. Conversely, \textbf{task-conditioned} approach optimize representations directly for task labels, the core representation learning is guided by the task objective. For example, PIBD \cite{zhang2024pibd} integrates information bottlenecks to explicitly learn prototypes associated with different risk levels and select relevant features. 
\section{Methods}
\label{sec:method}
\begin{figure*}[t]
  \centering
  \includegraphics[width=1.0\textwidth]{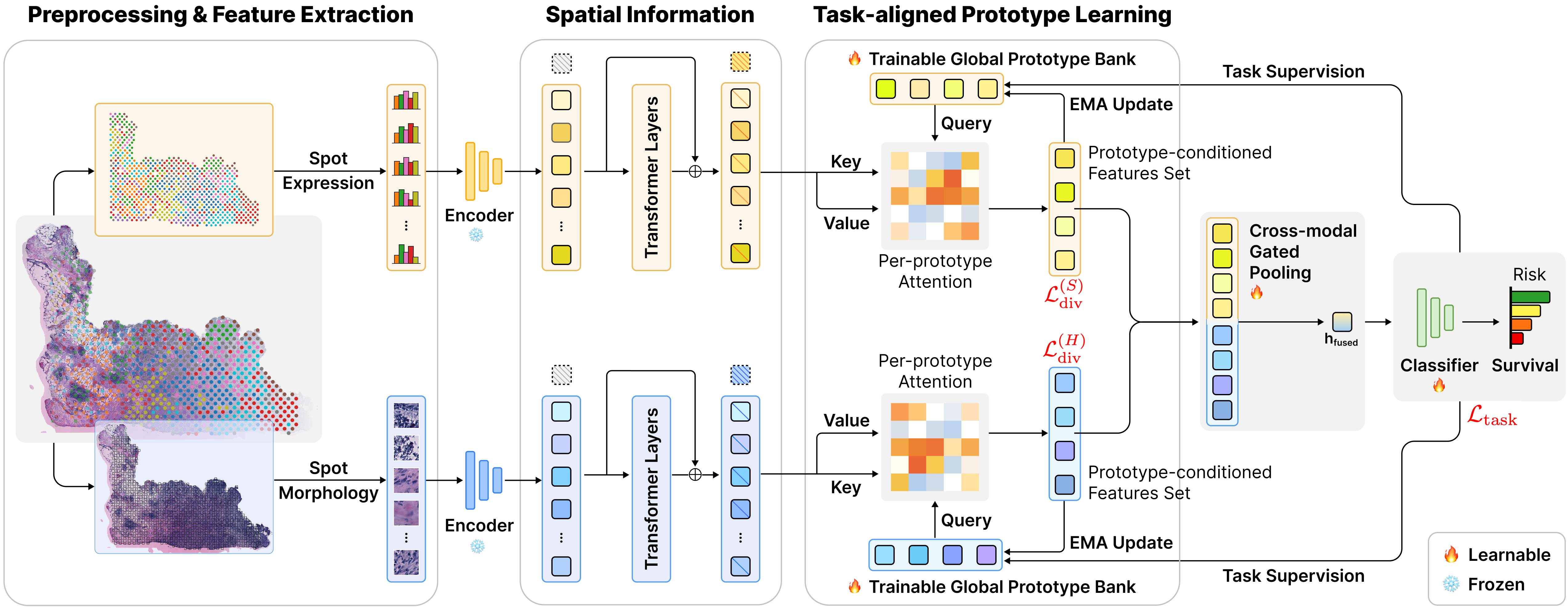}
  \caption{
    \textbf{Model Overview}. $\ours$ integrates co-registered whole-slide images (WSIs) and spatial transcriptomics (ST) via modality-specific encoders and adaptive prototype learning. Each modality learns an evolving prototype bank that captures morphology- or gene-level patterns. A fusion module performs cross-modal multiple-instance aggregation to produce a unified patient-level representation. The prognostic prediction task provides task-guided feedback that refines prototypes, enhancing discriminative and biological relevance. Prototype activations produce spatially interpretable maps linking histological regions and molecular signatures to survival risk.
    }
  \label{fig:fig1_overview}
\end{figure*}

We propose $\ours$, a novel framework integrating whole slide imaging data and spatial transcriptomics data to enable spatially-aware multimodal representation learning for survival prediction (Fig. \ref{fig:fig1_overview}). We first formulate the problem (Sec. \ref{sec:problem_formulation}), then explain the modality-specific encoders (Sec. \ref{sec:feature_extraction}). We then describe task-guided prototype learning (Sec. \ref{sec:prototype_learning}) and patient survival prediction (Sec. \ref{sec:final_loss_function}). Lastly, we introduce model interpretability via prototype annotation and risk prediction decomposition (Sec. \ref{sec:model_interpretability}).

\subsection{Problem Formulation}
\label{sec:problem_formulation}

In Spatial Transcriptomics (ST) datasets, each patient provides multiple tissue slides, each consists of two complementary modalities: \textbf{\textbf{Spatial transcriptomics (S):}} gene expression vectors at each spatially indexed round capture area (`spot'), reflecting local cellular states and transcriptional programs. \textbf{\textbf{Histology (H):}} a co-registered high resolution, e.g., gigapixel, whole slide images of the tissue section covering all the spots above. 

In a weakly supervised multi-modal MIL setting, each patient may contribute multiple slides, each slide $i$ forms a bag: $B_i = \{(\mathbf{x}_{ij}^{(H)}, \mathbf{x}_{ij}^{(S)})\}_{j=1}^{N_i}$, where spatially aligned histology patches $\mathbf{x}_{ij}^{(H)}$ and transcriptomic spots $\mathbf{x}_{ij}^{(S)}$ share slide-level labels $y_i$. The goal is to predict patient-level survival risk $y$ from these bags.

\subsection{Modality-specific Feature Extraction}
\label{sec:feature_extraction}
For each modality, we apply widely used specialized pre-trained encoders to generate modality-specific embeddings. For \textbf{spatial transcriptomics}, each spot expression profile $\mathbf{x}_{ij}^{(S)}$ is processed using scGPT~\cite{cui2024scgpt}, producing molecular embeddings $\mathbf{m}_{ij}^{(S)} \in \mathbb{R}^{512}$. For \textbf{histology}, each patch $\mathbf{x}_{ij}^{(H)}$ covering the co-registered spot is encoded using UNI2~\cite{chen2024uni}, generating morphological embeddings $\mathbf{m}_{ij}^{(H)} \in \mathbb{R}^{1536}$. 

To capture spatial dependencies between tissue locations within each modality, we leverage a Transformer-based spatial encoder inspired by TransMIL~\cite{shao2021transmil}. Embeddings $\mathbf{m}_{ij}^{(m)}$ from each modality ($m \in \{H,S\}$) are first transformed into $\mathbf{h}_{ij}^{(m)} \in \mathbb{R}^{512}$ with a linear projection. Following TransMIL, we apply Pyramid Positional Encoding (PPEG) and pass the tokens through two Transformer layers to model local spatial context, yielding spatially contextualized embeddings $\mathbf{T}^{(m)}$. We discard the deliberately prepended class token, thereby preserving the original sequence length. To retain both global context and local detail, we also intend to preserve the high-fidelity local information in the initial embedding sequence $\mathbf{h}_{ij}^{(m)}$. The final embeddings are computed as a weighted residual connection:
\begin{equation}
\tilde{\mathbf{h}}_{ij}^{(m)} = (1-\alpha) \text{LayerNorm}({\mathbf{h}}_{ij}^{(m)}) + \alpha {\mathbf{t}}_{ij}^{(m)}
\end{equation}
where $\mathbf{t}_{ij}^{(m)} \in \mathbf{T}^{(m)}$ and $\alpha\in (0,1)$. This strategy retains detailed local features from the modality-specific encoders while using global context as a refinement signal, thereby preserving fine-grained biological detail for the subsequent prototype learning stage. Details can be found in \textbf{Supplementary Material} Sec. \ref{sec:spatial_dependencies_supp}.

\subsection{Adaptive Multi-Level Mixture-of-Experts}
\label{sec:prototype_learning}
We hypothesize that each modality encodes distinct yet complementary prognosis-relevant representations, e.g., recurring pathological patterns and gene expression signature programs, and that integrating them improves predictive performance. Accordingly, we formulate the summarization of recurring pattern as an unsupervised objective and predictive aggregation as a supervised objective, and jointly optimize them so the learned representations align with survival prediction. Guided by the core MoE principles, we arrange experts sequentially in a multi-level hierarchy aligned with the two objectives and train the model end-to-end.

\subsubsection {\textbf{Modality-specific Prototype Expert}}
\label{sec:proto_conditioned_feature}

In each modality, we model the summarization of recurring patterns within a bag as learning attention-weighted means over homogeneous instance subsets, e.g., similar tissue structure and cellular shapes, in histology, and coordinated metabolic pathways and gene signatures in ST. Specifically, this process consists of 1.) dynamic instance selection via a routing mechanism and 2.) adaptive aggregation, by designated experts. We see both operations can be mediated by cross attention, whose scores jointly support selection, i.e., top-k selection, and aggregation, i.e., attention weighted mean. 

As such, for each modality \(m\in\{H,S\}\), we maintain learnable global prototypes \(\mathbf{P}^{(m)}=\{\mathbf{p}_k^{(m)}\}_{k=1}^{K_m}\) that act as specialized experts, each targeting a distinct recurring pattern. Each prototype is a trainable vector \(\mathbf{p}_k^{(m)}\in\mathbb{R}^{d_{\text{proto}}}\); in our model we set \(d_{\text{proto}}=512\). These modality-specific prototypes are expected to represent complementary patterns. (See global prototype training strategy at Sec. \ref{sec:global_prototype_learning})

Given a bag of input instances, an effective prototype is expected to attend to the most informative signals of the designated biological concept, tune out the irrelevant noise and yield prototype-conditioned features for subsequent prediction. Our model mimics this through a cross attention module followed by top-k selection. 

For each prototype $k$ in modality $m$ on slide $i$, we first compute the attention score (i.e., cosine similarity) between the prototype and every spatial instance (patch or spot) on the slide, forming a vector of scores for all $N_i$ instances. We retain only the top $k_{im}$ instances with the highest similarities $
\mathcal{I}_{ik}^{(m)} = \mathrm{TopK}_{k_{im}}(\mathbf{s}_{i,k,:}^{(m)})$. Specifically, the number of instances to keep $k_{im} = \min\left(\texttt{k}_{\min}, \lfloor N_i/K_m \rfloor\right)$ is set as the smaller of a predefined minimum threshold and the average number of instances per prototype, where \(k_{\min}\) is typically smaller than \(N_i/K_m\) since we intend to let the model learn only the strong signal. A sensitivity analysis of the number of selected instances in \textbf{Supplementary Material} Sec. \ref{sec:sensitivity_analysis}.

Once the top-k instances are selected, convert the similarities into normalized attention weights through softmax function:
\begin{equation}
\pi_{ikj}^{(m)} = \frac{\exp(s_{ikj}^{(m)})}{\sum_{j' \in \mathcal{I}_{ik}^{(m)}} \exp(s_{ikj'}^{(m)})}
\end{equation}

These weights ensure that instances with higher similarity have greater influence on the prototype update. Finally, we compute the prototype-conditioned feature set as the weighted average of its most relevant features:
\begin{equation}
\tilde{\mathbf{p}}_{ik}^{(m)} = \sum_{j \in \mathcal{I}_{ik}^{(m)}} \pi_{ikj}^{(m)} \tilde{\mathbf{h}}_{ij}^{(m)}
\end{equation}
Each element in the prototype-conditioned feature set represents a distinct biological concept (e.g., tumor regions or immune infiltration in histology, hypoxia or cell proliferation pathways in spatial transcriptomics), while automatically filtering out background and irrelevant information. This feature set is then used for the current prediction.

\subsubsection{Global Prototype Learning}
\label{sec:global_prototype_learning}

We define a set of global prototypes that act as dataset-wide descriptors shared across all bags. Each prototype is randomly initialized and intended to serve as the centroid of a recurring pattern. Prototypes are model parameters and are optimized jointly with the network using the survival prediction loss via backpropagation, enabling them to capture signals relevant to the supervised objective. 

In addition, to preserve method stability and avoid overfitting to the task loss, we also include an unsupervised objective that updates the prototypes in a data-driven manner, encouraging them to represent consistent, reusable patterns applicable across the training set. Specifically, we update the global prototype bank using an exponential moving average (EMA) of the training-set averaged prototype-conditioned feature set:
\begin{equation}
\mathbf{P}^{(m)} \leftarrow \beta \mathbf{P}^{(m)} + (1-\beta) \bar{\mathbf{P}}_{\text{epoch}}^{(m)}
\end{equation}
where $\beta\in (0,1)$. Here  $\bar{\mathbf{P}}_{\text{epoch}}^{(m)}$ is the mean of prototype-conditioned feature set of all the slides for modality $m$ computed at the end of each training epoch. This EMA update acts as an anchor for stability, which smooths prototype evolution, mitigates noisy fluctuations from batches, and steers the global prototypes toward cohort-level distributions while preserving alignment with the prediction task.

\subsubsection{Prototype diversity regularization}
\label{sec:prototype_diversity_regularization}

While the top-k selection strategy ensures that each prototype focuses on its own most relevant patches or spots, without additional regularization multiple prototypes might converge toward the same feature subspace. For example, all attending to tumor cores while ignoring immune regions or stromal tissue.
To encourage specialization, we apply a diversity penalty that discourages redundancy between prototypes within each modality:
\begin{equation}
\mathcal{L}_{\mathrm{div}}^{(m)} = \sum_{u \neq v}\left(\frac{\langle \mathbf{p}_u^{(m)}, \mathbf{p}_v^{(m)}\rangle}{\|\mathbf{p}_u^{(m)}\|_2 \|\mathbf{p}_v^{(m)}\|_2}\right)^2.
\end{equation}

\subsubsection{Cross-Modality Fusion via Gated Attention}
\label{sec:prototype_aggregation}

The final step fuses prototype-conditioned feature set from both modalities to a multimodal bag representation. We stack slide-specific prototypes from the histology and ST branches:
\begin{equation}
\mathcal{H}_i = [\tilde{\mathbf{P}}_i^{(H)}; \tilde{\mathbf{P}}_i^{(S)}] \in \mathbb{R}^{(K_H+K_S) \times d}.
\end{equation}

An input-dependent cross-modal fusion expert, implemented as an gated attention network, computes fusion weights $(\alpha_r)$ for each prototype. The final fused representation is the weighted sum of all experts:
\begin{equation}
\mathbf{h}_{\text{fused}} = \sum_{r=1}^{K_H+K_S} \alpha_r \mathbf{h}_r.
\end{equation}

In this final expert framework, we dynamically aggregate prototypical patterns from each modality, akin to integrating multiple sources of evidence, and automatically weight their contributions based on each slide’s context.

\subsection{Survival prediction}
\label{sec:final_loss_function}
Our model is trained end-to-end by optimizing a composite loss function that integrates task-specific supervision with interpretability-promoting regularization terms. Specifically, the final loss function is defined as:
\begin{equation}
\mathcal{L}_{\mathrm{final}} = \mathcal{L}_{\mathrm{task}} + \lambda_{\mathrm{div}}\left(\mathcal{L}_{\mathrm{div}}^{(H)} + \mathcal{L}_{\mathrm{div}}^{(S)}\right),
\end{equation}
where $\mathcal{L}_{\mathrm{task}}$ is the Negative Log-Likelihood (NLL) survival loss \cite{zadeh2020nll_loss}, and $\mathcal{L}_{\mathrm{div}}^{(m)}$ is the diversity penalty enforcing orthogonality among prototypes within each modality \( m \in \{H, S\} \), as defined earlier in Sec. \ref{sec:prototype_diversity_regularization}. The hyperparameter $\lambda_{\mathrm{div}}$ balances task performance and interpretability by controlling the strength of prototype diversity regularization.

\subsection{Model Interpretability} 
\label{sec:model_interpretability}

A key advantage of our prototype design is interpretability. We illustrate this with two complementary analyses.

\noindent\textbf{Prototype Characterization}
Each global prototype is dynamically adapted into a slide-specific representation, assigning cosine-similarity scores to all instance embeddings. Formally, for a bag or slide containing $N$ patches and $P$ prototypes, we compute the similarity (attention) matrix $\mathbf{S}\in\mathbb{R}^{P\times N}$. For each \textbf{histology prototype}, we select the highest-attention patches as representatives and spatially align them with pixel-level expert annotations (e.g., tumor, stroma, lymphoid, necrosis) to visually assess concordance. We then compute pairwise similarity between prototype attention scores and the pixel counts of each annotation category to quantify this concordance. For \textbf{ST prototypes}, we select top-attended spots as representative instances, merge highly similar prototypes, perform differential expression (DE) on each merged set, and then run over-representation analysis (ORA) of DE genes to derive functional annotations. Overall, this step links otherwise uninterpretable prototypes to biology-grounded tissue functions. More details can be found in 
\textbf{Supplementary Material} Sec. \ref{sec:model_interpretation}.

\noindent\textbf{Biological Decomposition of Risk Prediction} We visualize the attention weights from cross-modality prototype fusion (Sec.~\ref{sec:prototype_aggregation}) for each prototype alongside its functional annotation, revealing each prototype’s relative contribution to the sample-level representation. This provides an interpretable decomposition of patient-level risk.
\section{Experiments}
\label{sec:experiment}

\subsection{Datasets}
We evaluate our method on one public Triple Negative Breast Cancer (TNBC) dataset \cite{wang2024spatialtnbc}, which consists of 273 samples from 92 treatment-na\"ive patients at sampling (approximately 3 consecutive slides per sample). There are five different survival endpoints: \textit{Distant Relapse-Free Survival (DRFS)}, \textit{Relapse-Free Survival (RFS)}, \textit{Invasive Breast Cancer-Free Survival (IBCFS)}, \textit{Invasive Disease-Free Survival (IDFS)} and \textit{Overall Survival (OS)}. Additional details can be found in \textbf{Supplementary Material} Sec. \ref{sec:TNBC_dataset}.

\begin{table*}[t]
\begin{threeparttable}
  \caption{Survival prediction results C-index (mean$\pm$std) over 5 survival event metrics. Best performance in \textbf{bold}, second best \underline{underlined}.}
  \label{tab:full_results}
  \centering
  \begin{tabular}{c|l|ccccc}
    \toprule
    Modality & Methods & DRFS & RFS & IBFS & IDFS & OS \\
    \midrule

    \multirow{8}{*}{Unimodal}
      & CLAM (\textit{H}) & 0.596 $\pm$ 0.06 & 0.604 $\pm$ 0.04 & 0.550 $\pm$ 0.10 & 0.594 $\pm$ 0.08 & 0.582 $\pm$ 0.07 \\
      & TransMIL (\textit{H}) & 0.628 $\pm$ 0.05 & 0.628 $\pm$ 0.03 & 0.547 $\pm$ 0.10 & 0.610 $\pm$ 0.12 & 0.641 $\pm$ 0.05 \\
      & Patch-GCN (\textit{H}) & 0.611 $\pm$ 0.08 & 0.614 $\pm$ 0.05 & 0.551 $\pm$ 0.15 & 0.577 $\pm$ 0.15 & 0.615 $\pm$ 0.03 \\
      & PANTHER (\textit{H}) & 0.560 $\pm$ 0.09 & 0.573 $\pm$ 0.03 & 0.542 $\pm$ 0.17 & 0.557 $\pm$ 0.15 & 0.557 $\pm$ 0.05 \\
    \cmidrule(lr){2-7} 
      & CLAM (\textit{S}) & 0.538 $\pm$ 0.11 & 0.601 $\pm$ 0.04 & 0.476 $\pm$ 0.12 & 0.477 $\pm$ 0.09 & 0.537 $\pm$ 0.10 \\
      & TransMIL (\textit{S}) & 0.605 $\pm$ 0.08 & 0.542 $\pm$ 0.10 & 0.568 $\pm$ 0.05 & 0.587 $\pm$ 0.03 & 0.607 $\pm$ 0.08 \\
      & Patch-GCN (\textit{S}) & 0.620 $\pm$ 0.10 & 0.615 $\pm$ 0.11 & 0.610 $\pm$ 0.10 & 0.642 $\pm$ 0.06 & 0.617 $\pm$ 0.10 \\
      & PANTHER (\textit{S}) & 0.613 $\pm$ 0.08 & 0.557 $\pm$ 0.07 & 0.564 $\pm$ 0.05 & 0.560 $\pm$ 0.04 & 0.547 $\pm$ 0.02 \\
    \midrule

    \multirow{12}{*}{\makecell{Unimodal\\Fusion}}
      & CLAM (\textit{H+S}; c)  & 0.592 $\pm$ 0.05 & 0.600 $\pm$ 0.04 & 0.537 $\pm$ 0.11 & 0.600 $\pm$ 0.09 & 0.593 $\pm$ 0.04 \\
      & CLAM (\textit{H+S}; dot) & 0.587 $\pm$ 0.07 & 0.539 $\pm$ 0.02 & 0.568 $\pm$ 0.09 & 0.570 $\pm$ 0.04 & 0.569 $\pm$ 0.04 \\
      & CLAM (\textit{H+S}; m) & 0.588 $\pm$ 0.07 & 0.540 $\pm$ 0.03 & 0.568 $\pm$ 0.09 & 0.554 $\pm$ 0.04 & 0.586 $\pm$ 0.07 \\
      & TransMIL (\textit{H+S}; c) & 0.623 $\pm$ 0.05 & 0.630 $\pm$ 0.06 & 0.585 $\pm$ 0.12 & 0.640 $\pm$ 0.07 & \underline{0.647 $\pm$ 0.03} \\
      & TransMIL (\textit{H+S}; dot) & 0.628 $\pm$ 0.05 & 0.600 $\pm$ 0.04 & 0.578 $\pm$ 0.13 & 0.633 $\pm$ 0.07 & 0.641 $\pm$ 0.03 \\
      & TransMIL (\textit{H+S}; m) & \underline{0.636 $\pm$ 0.04} & 0.623 $\pm$ 0.06 & 0.591 $\pm$ 0.10 & 0.642 $\pm$ 0.07 & \textbf{0.648 $\pm$ 0.03} \\
      & Patch-GCN (\textit{H+S}; c) & 0.617 $\pm$ 0.08 & 0.610 $\pm$ 0.08 & 0.589 $\pm$ 0.14 & 0.633 $\pm$ 0.10 & 0.632 $\pm$ 0.10 \\
      & Patch-GCN (\textit{H+S}; dot) & 0.617 $\pm$ 0.08 & 0.614 $\pm$ 0.07 & 0.599 $\pm$ 0.11 & \underline{0.652 $\pm$ 0.06} & 0.613 $\pm$ 0.08 \\
      & Patch-GCN (\textit{H+S}; m) & 0.595 $\pm$ 0.06 & 0.596 $\pm$ 0.09 & 0.570 $\pm$ 0.12 & 0.628 $\pm$ 0.08 & 0.594 $\pm$ 0.06 \\
      & PANTHER (\textit{H+S}; c) & 0.577 $\pm$ 0.09 & 0.579 $\pm$ 0.04 & 0.553 $\pm$ 0.18 & 0.548 $\pm$ 0.15 & 0.556 $\pm$ 0.05 \\
      & PANTHER (\textit{H+S}; dot) & 0.592 $\pm$ 0.06 & 0.542 $\pm$ 0.08 & 0.547 $\pm$ 0.12 & 0.562 $\pm$ 0.14 & 0.530 $\pm$ 0.09 \\
      & PANTHER (\textit{H+S}; m) & 0.582 $\pm$ 0.06 & 0.547 $\pm$ 0.06 & 0.501 $\pm$ 0.10 & 0.534 $\pm$ 0.11 & 0.550 $\pm$ 0.04 \\
    \midrule

    \multirow{10}{*}{Multimodal}
      & PORPOISE (\textit{H+G}; c) & 0.527 $\pm$ 0.08 & 0.569 $\pm$ 0.08 & 0.569 $\pm$ 0.04 & 0.639 $\pm$ 0.09 & 0.508 $\pm$ 0.05 \\
      & PORPOISE (\textit{H+G}; bi) & 0.548 $\pm$ 0.07 & 0.568 $\pm$ 0.06 & 0.555 $\pm$ 0.05 & 0.639 $\pm$ 0.09 & 0.511 $\pm$ 0.04 \\
      & MCAT (\textit{H+G}) & 0.617 $\pm$ 0.06 & 0.617 $\pm$ 0.02 & 0.580 $\pm$ 0.09 & 0.633 $\pm$ 0.10 & 0.610 $\pm$ 0.05 \\
      & SurvPath (\textit{H+G}) & 0.580 $\pm$ 0.06 & 0.615 $\pm$ 0.04 & 0.590 $\pm$ 0.15 & 0.580 $\pm$ 0.15 & 0.622 $\pm$ 0.04 \\
      & PIBD (\textit{H+G}) & 0.584 $\pm$ 0.06 & 0.620 $\pm$ 0.07 & 0.570 $\pm$ 0.11 & 0.615 $\pm$ 0.08 & 0.584 $\pm$ 0.09 \\
      & ProSurv (\textit{H+G}; c)  & 0.546 $\pm$ 0.07 & 0.602 $\pm$ 0.05 & 0.600 $\pm$ 0.05 & 0.610 $\pm$ 0.11 & 0.541 $\pm$ 0.04 \\
      & ProSurv (\textit{H+G}; bi) & 0.591 $\pm$ 0.06 & 0.611 $\pm$ 0.04 & 0.600 $\pm$ 0.05 & 0.638 $\pm$ 0.10 & 0.569 $\pm$ 0.05 \\
    \cmidrule(lr){2-7}
      & \textbf{Ours} ($K_H$ = 12; $K_S$ = 8) & \textbf{0.637 $\pm$ 0.05} & \underline{0.645 $\pm$ 0.06} & \underline{0.610 $\pm$ 0.15} & 0.647 $\pm$ 0.09 & 0.621 $\pm$ 0.06 \\
      & \textbf{Ours} ($K_H$ = 32; $K_S$ = 24) & 0.631 $\pm$ 0.05 & \textbf{0.646 $\pm$ 0.07} & \textbf{0.615 $\pm$ 0.15} & \textbf{0.662 $\pm$ 0.08} & 0.627 $\pm$ 0.04 \\
    \bottomrule
  \end{tabular}
  
  \begin{tablenotes}
    \small
    \item We utilize (1) \textbf{different features} in different evaluations, \textbf{\textit{H}} histology images, \textbf{\textit{S}} spatial transcriptomics and \textbf{\textit{G}} pseudo-bulk gene expression, (2) \textbf{different fusion strategies} unimodal fusion, \textbf{\textit{c}} concatenation, \textbf{\textit{dot}} dot product, \textbf{\textit{m}} mean, \textbf{\textit{bi}} bilinear.
    \end{tablenotes}
    \end{threeparttable}
\end{table*}

\subsection{Baseline Methods}
We compare our method with state-of-the-art uni- and multimodal MIL frameworks spanning spatial awareness, prototype learning and multimodal fusion. Specifically, we consider (1) \textbf{Unimodal methods}: CLAM \cite{lu2021clam} is spatial-unaware and uses instance clustering-constrained attention, PatchGCN \cite{chen2021patchgcn} is spatial-aware and uses local neighbor information, TransMIL \cite{shao2021transmil} is spatial-aware and uses global correlation, PANTHER \cite{song2024panther} uses unsupervised prototype for bag representation; (2) \textbf{Late fusion of unimodal methods}: we apply concatenation, dot product and mean as late-fusion strategies on all unimodal backbones. (3) \textbf{Multimodal methods}: PORPOISE \cite{chen2021porpoise} integrates highly variable genes and pathology using late fusion, MCAT \cite{chen2021mcat} encodes transcriptomics to functional categories and uses co-attention for fusion. SurvPath \cite{jaume2023survpath} encodes transcriptomics as biological pathways and uses self-attention for fusion, PIBD \cite{zhang2024pibd} uses fully supervised prototypes for bag representation, and ProSurv \cite{LiuFen2025prosurv} uses modality-specific prototype banks for multimodal translation and fusion.

\begin{figure*}[!t]
  \centering
  \includegraphics[width=1.0\textwidth]{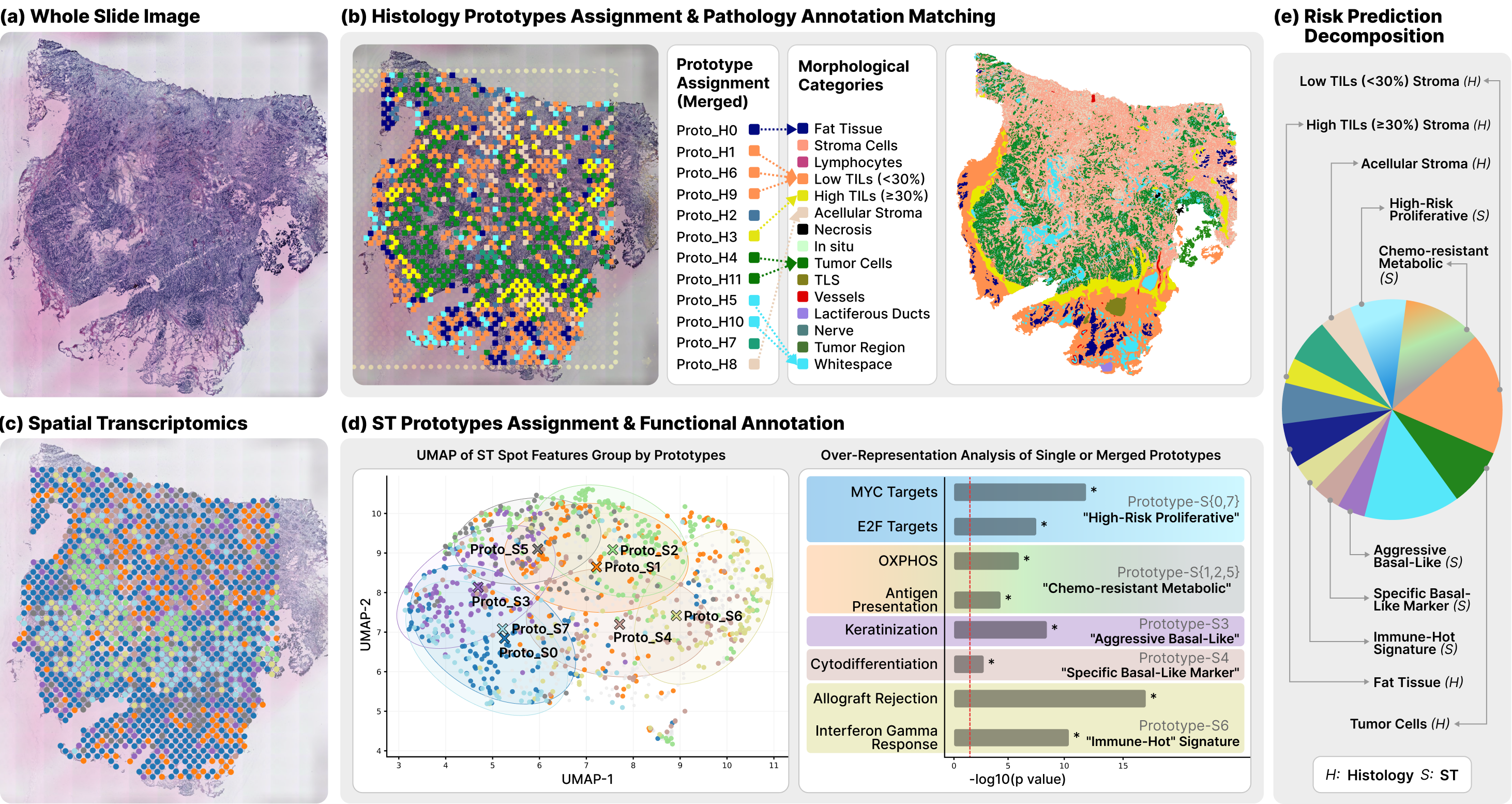}
  \caption{
    \textbf{Prototype Interpretation and Risk Decomposition}. (a) The whole slide image used in this sample. (b) The histology prototypes assignment and spatial location, we deliberately merged a few prototypes based on their spatial adjacency. Overall, they show high concordance with pathology annotation. (c) The ST prototypes assignment. (d) The UMAP of all ST spot features grouped by their prototypes and functional characterization in at the merged prototype level. (e) Per-prototype attention scores used by the model with interpretable semantics.}
  \label{fig:fig3_interpretation}
\end{figure*}

\subsection{Implementation Details}
\textbf{Data}: In this ST dataset, each spot is of $100\,\mu\mathrm{m}$ diameter which contains the full gene expression profile. The co-registered WSI is at $\sim 26\times$ magnification ($0.39\,\mu\mathrm{m}$/pixel). For spatial transcriptomics, we use all the valid spots in each sample for scGPT \cite{cui2024scgpt} embedding generation. For WSI, we first perform quality control with GrandQC \cite{Weng2024grandqc}, then crop out the non-overlapping patch of 256 $\times$ 256 pixels that are co-registered with ST spots, finally perform feature extraction with UNI2 \cite{chen2024uni}. For benchmarking methods that require bulk gene expression, we used pseudo-bulk expression by summing up all the spots. We encode pseudo-bulk expression to either functional categories or hallmarks pathways, concordant with the original paper. 

\noindent\textbf{Cross validation} We evaluate all models using 5-fold patient-level cross-validation to prevent data leakage. Following \cite{zadeh2020nll_loss}, survival time is discretized into quartile-based bins and the negative log-likelihood loss was used, the model predicts the interval-specific hazard for each bin.

\noindent\textbf{Model hyperparameters}: In our experiments, the weighted residual blending spatial-aware features with original features was set to $\alpha=$ 0.1; the min number of top-k selection \(k_{\min}\) was set to 60; the EMA updating magnitude is set to  $\beta=$ 0.95; the prototype diversity loss parameter is set to $\lambda_{\mathrm{div}}$ = 0.2. Additional details can be found in \textbf{Supplementary Material} Sec. \ref{sec:implementation_details}. 


\subsection{Ablation Study}
We ablate three key \textbf{model components}, \textit{spatial dependency modeling}, \textit{prototype learning} and the \textit{cross-modal prototype aggregation}, of our model to assess their contributions to the overall performance. We also ablate each data modality to assess their contributions.

\begin{table*}[!t] 
  \caption{Ablation study results include ablation of both model component and modality component. The best performance is highlighted in \textbf{bold}.}
  \label{tab:ablation_study}
  \centering
  \begin{tabular}{l|ccccc} 
    \toprule
    Model Variants & DRFS & RFS & IBFS & IDFS & OS \\
    \midrule
    w/o Self-attention feature blend & 0.612 $\pm$ 0.07 & 0.630 $\pm$ 0.08 & 0.582 $\pm$ 0.15 & 0.604 $\pm$ 0.12 & 0.632 $\pm$ 0.05 \\
    w/o prototype learning & 0.601 $\pm$ 0.05 & 0.628 $\pm$ 0.05 & 0.589 $\pm$ 0.12 & 0.604 $\pm$ 0.09 & 0.606 $\pm$ 0.04 \\
    w/o cross-modal attention & \textbf{0.653 $\pm$ 0.06} & 0.627 $\pm$ 0.07 & 0.535 $\pm$ 0.11 & 0.619 $\pm$ 0.11 & \textbf{0.639 $\pm$ 0.06} \\
    w/o ST Modality & 0.620 $\pm$ 0.06 & 0.627 $\pm$ 0.06 & 0.584 $\pm$ 0.10 & 0.632 $\pm$ 0.13 & 0.601 $\pm$ 0.03 \\
    w/o Histology Modality & 0.625 $\pm$ 0.08 & 0.586 $\pm$ 0.08 & 0.577 $\pm$ 0.10 & 0.591 $\pm$ 0.03 & 0.616 $\pm$ 0.07 \\
    \midrule
    Full Model ($K_H$ = 12; $K_S$ = 8) & 0.637 $\pm$ 0.05 & \textbf{0.645 $\pm$ 0.06} & \textbf{0.610 $\pm$ 0.15} & \textbf{0.647 $\pm$ 0.09} & 0.621 $\pm$ 0.06 \\
    \bottomrule
  \end{tabular}
\end{table*}
\section{Results}
\label{sec:results}
\subsection{Survival Prediction}
\label{sec:survival_prediction_results}

We benchmark the C-index based performance against nine uni- and multimodal state-of-the-art methods. We also test three late-fusion variants (concatenation, dot product, mean) combine with four unimodal baselines. Overall, $\ours$ shows superior or competitive performance across five survival endpoints (Table~\ref{tab:full_results}).

\noindent\textbf{$\boldsymbol{\ours}$ v.s. Unimodal Methods} In four endpoints (DRFS, RFS, IBFS and IDFS), $\ours$ showed superior performance than unimodal baselines with either histology or spatial transcriptomics modality, reflecting the gains from fusing complementary modalities. In addition, strong single-modality results from TransMIL and PatchGCN underscore the value of spatial information integration.

\noindent\textbf{$\boldsymbol{\ours}$ v.s. Unimodal Fusion} 
$\ours$ consistently outperforms baseline methods, except for OS, where TransMIL with averaged H and S features performs best. PatchGCN’s strong IDFS results highlight the relevance of spatial context. CLAM and PANTHER perform modestly; CLAM’s instance-level focus may miss heterogeneous signals, while PANTHER’s unsupervised prototypes may introduce noise by modeling weak or irrelevant patterns. Also, fusion is not uniformly superior, fusion of histology (H) and spatial transcriptomics (S) has method- and endpoint-dependent effects: it can improve performance (e.g., IBFS with CLAM; IDFS with PatchGCN), degrade it (e.g., CLAM on RFS; PatchGCN on DRFS), or approximate the unimodal average. These observations suggest the complex, sample-dependent cross-modal interactions, thus highlighting the need for adaptive, dynamic fusion strategies that can adjust to varying tissue context rather than relying on static fusion with fixed global alignment.

\noindent\textbf{$\boldsymbol{\ours}$ v.s. Multimodal Methods}
$\ours$ outperforms the selected multimodal baselines. Note that we used pseudo-bulk expression (G) as the genetic modality, which lacks the spatial granularity of in ST, this highlights the value of ST. Also, across benchmarks, different methods lead on different tasks, indicating task-dependent strengths.

\subsection{Interpretability}
\label{sec:survival_prediction_interpretibility}

\subsubsection{Sample-level Interpretability}
We demonstrate model interpretability in one sample ( ID:\textit{TNBC82-CN41-E1}) which was accurately predicted as medium-to-low risk in RFS prediction. We show 1.) morphological and biological interpretation of each prototype and 2.) decomposition of the risk prediction, see Fig \ref{fig:fig3_interpretation}. \noindent\textbf{Prototypes Interpretation}. For \underline{12 histology prototypes}, We select the top patches that each prototype expert attend to, and exclude the ambiguous ones. We observed high concordance between our prototype assignment with ground-truth pathological annotation categories. Specifically, our $prototype_{H1}$, $prototype_{H6}$ and $prototype_{H9}$ correlate with low TILs and stroma region, $prototype_{H3}$ align with high TILs region, $prototype_{H4}$ and $prototype_{H11}$ have high concordance with the tumor cell regions; while $prototype_{H0}$, $prototype_{H8}$, $prototype_{H5}$ and $prototype_{H10}$ correspond to fat, acellular stroma and white space. This demonstrates our prototypes can learn the intrinsic morphological patterns. Similarly, among the \underline{8 ST prototypes}, we identify intrinsic subgroups and merge prototypes based on their similarity as needed. We then performed differential gene expression analysis and over-representation analysis to obtain relevant biological characterization, such as MYC pathway \cite{bild2006mycpath}, E2F pathway \cite{bild2006mycpath}, oxidative phosphorylation (OXPHOS) \cite{elbotty2023oxphos}, keratinization \cite{rakha2025keratin}, allograft rejection \cite{oshi2022allograft}. \noindent\textbf{Risk prediction Decomposition}. We visualize the model’s attention over cross-modal prototypes and annotate key prototypes with interpretable semantic labels (e.g., pathological categories and biological pathways). Many attended prototypes accurately capture survival-relevant patterns, such as tumor cells, high/low tumor-infiltrating lymphocytes, immune-hot gene expression and high-risk proliferation, etc. 

\subsubsection{Cohort-level Interpretability}
We demonstrate model interpretability at the cohort level by evaluating prototype consistency and modality-oriented patient stratification among the whole cohort. textbf{Prototype Consistency} In the pathology modality, the mean adjusted Kappa score across all matched instances achieved $0.68 \pm 0.05$ across five endpoints, especially, tumor (0.85), TIL (0.83), and necrosis (0.78). In the ST modality, prototypes consistently mapped to key biological categories with high prevalence ($>60\%$), such as immune activation ($p < 10^{-12}$), ECM remodeling ($p < 10^{-10}$), and genome integrity ($p < 10^{-9}$). These signals are highly prognostic in breast cancer, underscoring the effective representation learning of our within-modality experts. Note that, the unmatched prototypes may represent novel signals that lie beyond current interpretation, such as fine-grained pathological semantics, topological features, or biological pathways not captured in standard gene set enrichment dictionaries. These signals hold the potential for new hypothesis generation and warrant further investigation. \textbf{Modality-Oriented Patient Stratification} Our model enables a modality-oriented decomposition of risk prediction. By aggregating attention scores from each modality, we can decompose patient risk predictions into specific profiles, for instance, identifying cases where the ST modality contribution outweighs H\&E, or vice versa. Such \textit{histo- vs. ST-dominant} patient profiles highlight a non-linear, sample-dependent relationship between modalities and demonstrate the advantage of our adaptive fusion expert. This decomposition may suggest novel stratification with the potential to refine personalized clinical treatment and justifies further exploration in future studies.

\subsection{Ablation Study and Sensitivity Analysis}

\label{sec:results_ablation}
In ablations of architectural components, the full model performs best on RFS, IBFS and IDFS. However, for DRFS and OS, replacing cross-modal aggregation with averaging improves performance. We attribute this to task-specific biases that penalize model complexity, thereby favoring simpler integration. For modalities, integrating both modalities consistently yields the best performance, evidencing complementary information essential for prediction. The results are shown in Table \ref{tab:ablation_study}. We further ablate three architectural components within each modality to assess modality-specific contributions. The full model achieves the best performance, indicating the essential contribution of each component. Detailed results are shown in \textbf{Supplementary Material} Sec. \ref{sec:further_ablation}. We also conducted a sensitivity analysis over (1) top-k patch selection for prototype updating, (2) prototypes per modality, and (3) gradient-accumulation steps; our performances remained comparable, which indicates that our model is robust to hyperparameter selection. The results are shown in \textbf{Supplementary Material} Sec. \ref{sec:sensitivity_analysis}.
\section{Conclusions and Limitations}
\label{sec:conclusion}

We present $\ours$, a multimodal and interpretable end-to-end MIL framework that is tailored to co-registered spatial transcriptomics and whole slide images to learn spatially informed prognostic representations. We specifically designed a hierarchical, multi-level expert architecture that combines within-modality, task-guided prototype experts with a dedicated cross-modal fusion expert to address intra-modal redundancy and complex cross-modal interactions.

$\ours$ represents a proof of concept for bridging this emerging data regime with clinical needs. Predictive performance and external validation are necessarily constrained by the limited availability of clinical-ready paired ST–WSI cohorts, due to high processing costs and long turnaround times. To our knowledge, this TNBC cohort is currently the only public dataset that supports end-to-end training of deep learning models by jointly providing adequate sample size, high-resolution WSIs, and relevant prognostic endpoints. Additionally, our end-to-end optimization can be further improved; more efficient and flexible training, such as step-wise training that first stabilizes prototype discovery and then fine-tunes the classifier, may further improve performance and convergence. Despite these limitations, the observed results align with the model’s design intent and support its central contribution of adaptive, interpretable fusion. As paired ST-profiled cohorts continue to expand in scale and quality, we expect $\ours$ to provide a foundation for future work integrating high-resolution spatial omics and computational pathology at scale.

\section{Code Availability}
\label{sec:code_availability}

\begin{sloppypar}
The project source code for training and benchmarking is available at \url{https://github.com/liulihe954/PathoSpatial}.
\end{sloppypar}
{
    \small
    \bibliographystyle{ieeenat_fullname}
    \bibliography{main}
}
\clearpage
\setcounter{page}{1}
\maketitlesupplementary
\setcounter{equation}{0}
\renewcommand{\theequation}{S\arabic{equation}}
\setcounter{table}{0}
\renewcommand{\thetable}{S\arabic{table}}
\setcounter{figure}{0}
\renewcommand{\thefigure}{S\arabic{figure}}

\section{Dataset Description}
\label{sec:TNBC_dataset}

\subsection{Sample Collection and Survival Endpoints}
We used public data from a retrospective study on early-stage TNBC patients (ER/PR/HER2-negative) treated at the Institut Jules Bordet between 2000 and 2016 \cite{wang2024spatialtnbc}. TNBC patients were selected based on their negative status for estrogen (ER) and progesterone (PR) receptors and the absence of HER2 amplification. All patients received initial surgery before getting adjuvant chemotherapy or radiotherapy. For each patient, a frozen surgical tumor tissue sample stored at $-80^\circ \text{C}$ was collected from the institutional tissue bank with their prior consent. Samples were required to have a tumor cellularity of greater than 15\%. 

This study tracked five standard survival endpoints based on STEEP criteria for Breast Cancer (BC), with a required follow-up of about five years, censored at ten years, and the clinical outcomes were defined following STEEP Version 2.0 criteria. Five standardized endpoints were reported as follows: 

\begin{itemize}
    \item \textbf{Distant relapse-free survival (DRFS)}:  time interval from the date of diagnosis to death related or not to BC or any distant recurrence.
    
    \item \textbf{Recurrence-free survival (RFS)}: time interval from the date of diagnosis to death related or not to BC, or any recurrence (distant or locoregional), excluding contralateral BC.
    
    \item \textbf{Invasive breast cancer-free survival (IBCFS)}: time interval from the date of diagnosis to death related or not to BC, or any recurrence (distant or locoregional), including contralateral BC.
    
    \item \textbf{Invasive disease-free survival (IDFS)}: time interval from the date of diagnosis to death related or not to BC, or any recurrence (distant or locoregional), including contralateral BC and second primary non-BC related invasive cancer.
    
    \item \textbf{Overall survival (OS)}: time interval from the date of diagnosis to death related or not to BC.
\end{itemize}

\subsection{Imaging and Sequencing}
Fresh-frozen tissue sections ($16\,\mu\mathrm{m}$) were placed onto ST microarrays consisting of 1934 spatially barcoded spots (100~$\mu$m diameter, 150~$\mu$m center-to-center spacing) printed on subarrays (6.2~mm~$\times$~6.4~mm; six subarrays per slide), each spot containing $\sim 200$~million uniquely barcoded oligonucleotides with poly-T$_{20}$VN capture regions. Sections were H\&E-stained and whole-slide images were acquired at $20\times$ magnification on a Zeiss AxioImager 2Z microscope using the Metafer Slide Scanning System, with image processing performed in VSlide. After imaging, coverslips and mounting medium were removed by aqueous and ethanol washes before downstream permeabilization and cDNA synthesis. Subsequently, tissues were permeabilized, subjected to on-slide reverse transcription, and cDNA was recovered for standard ST library preparation (second-strand synthesis, \textit{in vitro} transcription, adapter ligation, and PCR). Libraries were quality-controlled and sequenced on an Illumina NextSeq~500 (v2) at $\sim 100$~million paired-end reads per tissue section. After quality control and filtering, 14{,}019 genes that were commonly detected across all spots and all samples were retained for downstream analyses.

\subsection{Pathological Annotations}
For each patient, a total of fifteen histomorphological categories were available for one of the high resolution WSI slides. The annotations consist of both manual annotations from breast pathologists, and single-cell level refinement for three categories tumor cells, stroma cells and lymphocytes using automatic detection algorithms from QuPath. The available annotation categories available were recorded as follows:

\begin{itemize}
    \item \textbf{Tumor cells}: Tumor cells (QuPath)
    
    \item \textbf{Stroma cells}: Main non-immune cellular component of the stroma (QuPath)
    
    \item \textbf{Lymphocytes}: Lymphocytes infiltrating the tumor but also lymphocytes distant from the tumor edge (QuPath)

    \item \textbf{Tumor region}: Area with majority of tumor cells (manual)
    
    \item \textbf{Low TILs stroma}: Stroma including acellular stroma, stroma cells with estimated fewer than 30\% lymphocytes in the area (manual)
    
    \item \textbf{High TILs stroma}: Stroma including acellular stroma, stroma cells with estimated greater than or equal to 30\% lymphocytes in the area (manual)

    \item \textbf{Acellular stroma}: Stroma without cells enriched by collagen fibers (manual)
    \item \textbf{Fat tissue}: Fat tissue (manual)
    \item \textbf{Necrosis}: Necrosis which could be infiltrated by immune cells (manual)
    \item \textbf{In situ}: Carcinoma in situ (manual)
    \item \textbf{Tertiary Lymphoid Structures}: Dense cellular aggregate (GC) or lymphoid aggregates and lymphoid follicles without GC and consisted of CD20-positive B zones with CD3-positive T zone aggregates (manual)

    \item \textbf{Vessels}: Vessels (manual)
    \item \textbf{Lactiferous ducts}: Normal ductal glands including those with hyperplasia (manual)
    \item \textbf{Nerves}: Nerve (manual)
    \item \textbf{Heterologous element}: Ectopic structures as bone, teeth, etc. (manual)
    
\end{itemize}





\section{Implementation Details}
\label{sec:implementation_details}

\subsection{Model Training}

All models were developed and trained with Python version 3.10.18 and PyTorch version 2.2.0+cu121 using NVIDIA A100 GPUs, with the batch size set to 1 and the random seed fixed at 1. The Adam optimizer is employed for model optimization with a constant learning rate of 1e-4 and a weight decay of 5e-5. The dropout rate was set to 0.25 for all models whenever feasible. Each model was trained for up to 50 epochs with early stopping, using a patience of 10 epochs. We report the averaged concordance index (mean C-index over the best epoch and best $\pm$ 1 epochs) to assess how well the model ranks patient risk in the validation set.

\subsection{Benchmarking}

We implemented all benchmarking methods with hyperparameters largely following the defaults provided in the original publications, and highlight the modified and user-determined parameters below:

\begin{itemize}
    \item \textbf{CLAM}: We set \textit{instance\_eval} to be \textit{True} to include instance level clustering loss.
    
    \item \textbf{TransMIL}: We added a drop out rate of 0.25 in the first fully connected layer in addition to the original methods to enable fair comparison.
    
    \item \textbf{PatchGCN}: We set \textit{radius} to 8 for graph construction in both histology and spatial transcriptomics modalities. The number of graph convolutional networks \textit{num\_gcn\_layers} was set to 4 in both modalities.
    
    \item \textbf{PANTHER}: We set \textit{n\_proto} to 16 in each modality, corresponding to 16 mixture components (prototypical distributions) in the Gaussian mixture model. We set \textit{out} to \textit{allcat}, meaning the final slide representation is formed by concatenating all the descriptors for all components.
    
    \item \textbf{PORPOISE}: We retained 2,000 genes with the highest median absolute deviation (MAD) across samples in the pseudo-bulk expression matrix to represent the genomic feature profile of each sample. We set \textit{size\_arg} to \textit{small} to use a smaller version of the genomic SNN.
    \item \textbf{MCAT}: We encode pseudo-bulk expression to 50 hallmarks pathways and construct genomic SNN using varying size of each pathway. The performance results shown in the results table \ref{tab:full_results} was using bilinear fusion.
    \item \textbf{SurvPath}: We encode pseudo-bulk expression to 50 hallmarks pathways and construct genomic SNN using varying size of each pathway.
    \item \textbf{PIBD}: We encode pseudo-bulk expression to 50 hallmarks pathways and construct genomic SNN using varying size of each pathway. We adjust \textit{num\_patches} to 2000 according to the number of spot/patch per slide in our dataset. We set both \textit{ratio\_omics} and \textit{ratio\_wsi} to be 0.5 when determining the top-k value in each modality.
    \item \textbf{ProSurv}: We retained 2,000 genes with the highest median absolute deviation (MAD) across samples in the pseudo-bulk expression matrix to represent the genomic feature profile of each sample.
\end{itemize}
\subsection{Loss Functions}
\textbf{Survival Loss}: We employ the Negative Log-Likelihood (NLL) survival loss as the primary prediction loss. We discretize continuous time into a set of non-overlapping intervals determined by the quartiles of survival times in the training set. For a patient $i$, let $y_i$ denote the discrete time bin index and $c_i \in \{0, 1\}$ denote the censorship status, where $c_i=1$ indicates censorship (event not observed) and $c_i=0$ indicates an observed event.

The model outputs a hazard probability $h_j$ for each time interval $j$, representing the conditional probability that the event occurs in interval $j$ given that the patient has survived up to interval $j-1$. The survival probability $S_j$, representing the probability that the patient survives past interval $j$, is defined recursively as the cumulative product of non-event probabilities:
\begin{equation}
    S_j = \prod_{k=1}^{j} (1 - h_k)
\end{equation}
where $S_0 = 1$.

The loss function aims to maximize the likelihood of the observed data. For an uncensored patient who experiences the event at time interval $y_i$, the likelihood is defined as the probability of surviving up to $y_i-1$ and experiencing the event at $y_i$. For a censored patient known to survive up to interval $y_i$, the likelihood is simply the survival probability $S_{y_i}$. The formulation is defined as:

\begin{equation}
    \mathcal{L}_{uncensored} = - \sum_{i: c_i=0} \log(S_{y_i-1} \cdot h_{y_i})
\end{equation}

\begin{equation}
    \mathcal{L}_{censored} = - \sum_{i: c_i=1} \log(S_{y_i})
\end{equation}

Combining these terms, the total Negative Log-Likelihood loss is:
\begin{equation}
    \mathcal{L}_{NLL} = (1-\alpha)\mathcal{L}_{total} + \alpha \mathcal{L}_{uncensored}
\end{equation}
where $\mathcal{L}_{total} = \mathcal{L}_{uncensored} + \mathcal{L}_{censored}$ and $\alpha$ is a weighting hyperparameter. In our experiments, we set $\alpha = 0.4$, where we encourage the model to focus on learning from the exact event times provided by uncensored samples.

\noindent \textbf{Prototype Diversity Loss}: 
To ensure the learned prototypes capture distinct features, we apply a diversity loss that encourages orthogonality among the learned prototypes. Let $\mathbf{P} \in \mathbb{R}^{K \times d}$ contain the $K$ prototype vectors as rows. We first $\ell_2$-normalize each row to obtain $\hat{\mathbf{P}}$. The cosine similarity (Gram) matrix is then $\mathbf{G} = \hat{\mathbf{P}} \hat{\mathbf{P}}^\top$. The diversity loss is defined as the mean squared error between $\mathbf{G}$ and the identity matrix on off-diagonal elements:

\begin{equation}
\mathcal{L}_{\text{div}}(\mathbf{P}) = \mathbb{E}_{i,j} \left[ (\mathbf{G} - \mathbf{I})_ {ij}^2 \right] = \frac{1}{K^2} \sum_{i,j} (\mathbf{G}_{ij} - \delta_{ij})^2,
\end{equation}
This loss is zero if and only if the prototypes are mutually orthogonal (up to scaling) and increases with their pairwise cosine similarity. We apply this diversity loss in each of the modalities separately.

\section{Methods}
\subsection{Spatial Context Modeling}
\label{sec:spatial_dependencies_supp}

To capture long-range dependencies while preserving local morphological details, we implement a modified TransMIL encoder. This process is applied independently to each modality $m \in \{H,S\}$. For clarity, we drop the modality superscript $(m)$ and sample index $i$ in this section, describing the operations on a sequence of patches for a single slide. Let $\mathbf{M} \in \mathbb{R}^{N \times D_{input}}$ denote the input bag of features (corresponding to the collection of embeddings $\mathbf{m}_{j}$), where $N$ is the number of patches.

\textbf{Feature Projection and Dimension Alignment}: Both histology features and spatial transcriptomics features are projected to a common latent dimension $D=512$ to ensure cross-modal alignment. Specifically, the input features $\mathbf{M}$ are transformed using a fully connected layer followed by a ReLU activation and Dropout ($p=0.25$). This yields the initial local feature sequence $\mathbf{H} \in \mathbb{R}^{N \times D}$, consisting of the embeddings $\mathbf{h}_{j}$ described in the main text.

\textbf{Sequence Padding and Tokenization}: The Pyramid Positional Encoding Generator (PPEG) module in TransMIL requires input tokens to be arranged in a 2D grid. Since $N$ varies and is rarely a perfect square, we perform dynamic padding following TransMIL implementation:
\begin{enumerate}
    \item We calculate the spatial dimensions of the closest enclosing square grid, $S = \lceil \sqrt{N} \rceil$.
    \item The sequence $\mathbf{H}$ is padded to length $S^2$ by repeating the initial patches of the sequence until the length requirement is met. This padding strategy ensures that the PPEG module operates on valid feature information rather than zero-tokens.
    \item A learnable classification token ($\mathbf{h}_{cls}$) is prepended to the sequence.
\end{enumerate}

\textbf{Spatial Encoding Layers}: The sequence passes through a stack comprising a first Transformer layer, the PPEG module, and a second Transformer layer. We insert the PPEG between the transformer layers to leverage the inductive bias of image-like structures:
\begin{align}
    \mathbf{Z}_{0} &= [\mathbf{h}_{cls} \mathbin{\Vert} \mathbf{H}_{padded}] \\
    \mathbf{Z}_{1} &= \text{TransLayer}_1(\mathbf{Z}_{0}) \\
    \mathbf{Z}_{pos} &= \text{PPEG}(\mathbf{Z}_{1}, H=S, W=S) \\
    \mathbf{Z}_{out} &= \text{TransLayer}_2(\mathbf{Z}_{pos})
\end{align}
Here, $\mathbin{\Vert}$ denotes concatenation. The PPEG module reshapes the sequence into an $S \times S$ grid, applies depth-wise separable convolutions, and flattens the result back to a sequence.

\textbf{Feature Recovery and Weighted Residual}: Post-encoding, the classification token is discarded, and the padded tail is truncated to restore the original sequence length $N$, resulting in the context-aware features $\mathbf{T} \in \mathbb{R}^{N \times D}$.

To ensure the final embeddings retain high-fidelity local information while benefiting from global context, we employ the weighted residual connection. Both the original projected features $\mathbf{H}$ and the transformer output $\mathbf{T}$ undergo Layer Normalization. The final embedding feature set $\tilde{\mathbf{H}}$ is computed as:
\begin{equation}
    \tilde{\mathbf{H}} = (1-\alpha) \cdot \text{LayerNorm}(\mathbf{H}) + \alpha \cdot \text{LayerNorm}(\mathbf{T})
\end{equation}
Consistent with our experiments, we set $\alpha = 0.1$. This weighting prioritizes the preservation of the original feature manifold $\mathbf{H}$ while using $\mathbf{T}$ as a spatial refinement signal.

\subsection{Prototype-Conditioned Feature Set}
\label{sec:proto_conditioned_feature_sup}

This section details the generation of the prototype-conditioned feature set described in Sec. \ref{sec:proto_conditioned_feature}. This process is applied independently to each modality, we drop the modality superscript $(m)$ and sample index $i$ in this section for clarity. Let $\tilde{\mathbf{H}} \in \mathbb{R}^{N \times D}$ be the spatially contextualized feature bag and $\mathbf{P} \in \mathbb{R}^{K \times D}$ be the learnable global prototype bank.

Within each forward pass, we compute the bag-level \emph{prototype-conditioned feature set} $\tilde{\mathbf{P}} \in \mathbb{R}^{K \times D}$ using a specialized Multi-Head Attention mechanism. We treat global prototypes as queries and instance tokens as keys and values.

First, the prototypes and instance features are projected into the attention subspace:
\begin{equation}
\mathbf{Q} = \mathbf{P}\mathbf{W}_Q, \quad \mathbf{K} = \tilde{\mathbf{H}}\mathbf{W}_K
\end{equation}
where $\mathbf{W}_Q$ and $\mathbf{W}_K$ are learnable projection matrices. To strictly measure directional alignment rather than magnitude, we compute the similarity matrix $\mathbf{S} \in \mathbb{R}^{K \times N}$ using Cosine Similarity:
\begin{equation}
S_{kj} = \frac{\mathbf{q}_k \cdot \mathbf{k}_j}{\|\mathbf{q}_k\|_2 \|\mathbf{k}_j\|_2}
\end{equation}
where $\mathbf{q}_k$ and $\mathbf{k}_j$ denote the rows of $\mathbf{Q}$ and $\mathbf{K}$, respectively.

To filter out noise and enforce concept specialization, we employ a \emph{sparse attention} mechanism. For each prototype $k$, we select only the top subset of instances. Let $\mathcal{I}_k = \mathrm{TopK}_{M}(\mathbf{S}_{k,:})$ be the indices of the $M$ largest similarity scores for the $k$-th prototype. We set a fixed instance count of $M=60$ to capture sufficient biological variation while maintaining specificity. For small slides, this threshold is dynamically adjusted:
\begin{equation}
M = \min\left( 60, \left\lfloor \frac{N}{K} \right\rfloor \right)
\end{equation}
This operation is applied row-wise, ensuring that each prototype focuses on its most relevant instances. Note that while the attention is sparse (ignoring most patches), a single highly informative instance may be selected by multiple prototypes if it exhibits compound biological traits.

We then form a \emph{masked} logit matrix $\mathbf{A}$ by assigning $-\infty$ to all unselected entries:
\begin{equation}
A_{k j} =
\begin{cases}
S_{k j}, & j \in \mathcal{I}_k \\
-\infty, & j \notin \mathcal{I}_k
\end{cases}
\end{equation}
The final attention weights $\mathbf{\Pi} \in \mathbb{R}^{K \times N}$ are obtained via a row-wise softmax, ensuring probability mass is distributed \emph{only} over the selected top-k patches:
\begin{equation}
\Pi_{k j} = \frac{\exp(A_{k j})}{\sum_{j' \in \mathcal{I}_k} \exp(A_{k j'})}
\end{equation}
Finally, the prototype-conditioned feature $\tilde{\mathbf{p}}_{k}$ is computed as the weighted average of the \emph{original} spatial features $\tilde{\mathbf{H}}$ (prior to projection or normalization):
\begin{equation}
\tilde{\mathbf{p}}_{k} = \sum_{j \in \mathcal{I}_k} \Pi_{k j} \tilde{\mathbf{h}}_{j}
\end{equation}
Stacking these vectors yields the final modality-specific feature set $\tilde{\mathbf{P}} = [\tilde{\mathbf{p}}_{1}, \dots, \tilde{\mathbf{p}}_{K}]^\top$.

\subsection{Cross-Modality Fusion via Gated Attention}
\label{sec:crossmodal_fusion_supp}

To synthesize a unified patient representation from the distinct biological signals captured by the prototypes, we employ a Gated Attention mechanism. This approach allows the model to dynamically assign importance to each prototype based on the specific context of the slide, effectively functioning as an expert to softly select risk-relevant features. For clarity, we drop the sample index $i$ in this section.

\textbf{Feature Stacking}: First, we aggregate the prototype-conditioned feature sets from both the Histology ($H$) and Spatial Transcriptomics ($S$) branches into a single candidate pool. Let $\tilde{\mathbf{P}}^{(H)} \in \mathbb{R}^{K_H \times D}$ and $\tilde{\mathbf{P}}^{(S)} \in \mathbb{R}^{K_S \times D}$ be the output features from the previous stage (with $D=512$). We stack these along the instance dimension to form the multimodal bag representation $\mathcal{H} \in \mathbb{R}^{R \times D}$, where $R = K_H + K_S$ is the total number of prototypes from two modalities:
\begin{equation}
\mathcal{H} = \left[ \tilde{\mathbf{P}}^{(H)} \mathbin{;} \tilde{\mathbf{P}}^{(S)} \right]
\end{equation}
where $\mathbin{;}$ denotes vertical concatenation. Let $\mathbf{h}_r \in \mathbb{R}^D$ denote the $r$-th row of $\mathcal{H}$ (i.e., the feature vector of the $r$-th item in the joint prototypes).

\textbf{Gated Attention Mechanism}: To compute the attention scores, we map the input features into a latent attention space using two separate linear projections followed by non-linear activations. Specifically, we employ the Gated Attention mechanism, which uses a sigmoid activation to gate the output of a tanh activation. This allows the model to learn complex non-linear relations among instances.

For each prototype feature $\mathbf{h}_r$, we compute the unnormalized attention score $s_r$ as follows:
\begin{equation}
s_r = \mathbf{w}_c^\top \left( \tanh\left(\mathbf{W}_a \mathbf{h}_r^\top\right) \odot \sigma\left(\mathbf{W}_b \mathbf{h}_r^\top\right) \right)
\end{equation}
where:
\begin{itemize}
    \item $\mathbf{W}_a, \mathbf{W}_b \in \mathbb{R}^{D' \times D}$ are learnable weight matrices projecting the input to the internal attention dimension $D'=256$.
    \item $\mathbf{w}_c \in \mathbb{R}^{D'}$ is a learnable weight vector projecting the gated features to a scalar score.
    \item $\odot$ denotes element-wise multiplication.
    \item $\sigma(\cdot)$ is the sigmoid activation function.
\end{itemize}
Dropout ($p=0.25$) is applied after the $\tanh$ and $\sigma$ activations during training to prevent overfitting.

\textbf{Bag Feature Aggregation}: The attention scores are normalized using a softmax function to obtain a probability distribution over the prototypes:
\begin{equation}
\alpha_r = \frac{\exp(s_r)}{\sum_{k=1}^{R} \exp(s_k)}
\end{equation}
The final fused representation $\mathbf{h}_{\text{fused}} \in \mathbb{R}^D$ is computed as the sum of the prototype features weighted by their respective attention scores:
\begin{equation}
\mathbf{h}_{\text{fused}} = \sum_{r=1}^{R} \alpha_r \mathbf{h}_r
\end{equation}
This vector $\mathbf{h}_{\text{fused}}$ serves as a compact, multimodal summary of the patient's slide, prioritizing the most prognostically relevant morphological and molecular patterns.

\begin{table*}[!t] 
  \caption{Ablation study results of model components with only histology (H). The best performance is highlighted in \textbf{bold}.}
  \label{tab:ablation_study_only_h}
  \centering
  \begin{tabular}{l|ccccc} 
    \toprule
    Model Variants & DRFS & RFS & IBFS & IDFS & OS \\
    \midrule
    w/o Self-attention feature blend & 0.604 $\pm$ 0.09 & 0.580 $\pm$ 0.08 & 0.576 $\pm$ 0.09 & 0.593 $\pm$ 0.04 & 0.600 $\pm$ 0.09 \\
    w/o prototype learning & 0.595 $\pm$ 0.05 & 0.567 $\pm$ 0.11 & 0.560 $\pm$ 0.07 & 0.589 $\pm$ 0.03 & 0.594 $\pm$ 0.08 \\
    w/o gated attention aggregation & 0.610 $\pm$ 0.08 & 0.558 $\pm$ 0.08 & 0.555 $\pm$ 0.09 & 0.601 $\pm$ 0.03 & \textbf{0.603 $\pm$ 0.07} \\
    \midrule
    Full Model (Only H; $K_H$ = 12; $K_S$ = 8) & \textbf{0.620 $\pm$ 0.06} & \textbf{0.627 $\pm$ 0.06} & \textbf{0.584 $\pm$ 0.10} & \textbf{0.632 $\pm$ 0.13} & 0.601 $\pm$ 0.03 \\      
    \bottomrule
  \end{tabular}
\end{table*}


\begin{table*}[!t] 
  \caption{Ablation study results of model components with only spatial transcriptomics (S). The best performance is highlighted in \textbf{bold}.}
  \label{tab:ablation_study_only_g}
  \centering
  \begin{tabular}{l|ccccc} 
    \toprule
    Model Variants & DRFS & RFS & IBFS & IDFS & OS \\
    \midrule
    w/o Self-attention feature blend & 0.606 $\pm$ 0.07 & \textbf{0.589 $\pm$ 0.06} & 0.547 $\pm$ 0.10 & 0.586 $\pm$ 0.12 & 0.602 $\pm$ 0.06 \\
    w/o prototype learning & 0.588 $\pm$ 0.05 & 0.579 $\pm$ 0.08 & 0.568 $\pm$ 0.10 & 0.560 $\pm$ 0.15 & 0.590 $\pm$ 0.04 \\
    w/o gated attention aggregation & 0.610 $\pm$ 0.06 & 0.581 $\pm$ 0.08 & 0.571 $\pm$ 0.07 & 0.589 $\pm$ 0.12 &  0.600 $\pm$ 0.04 \\
    \midrule
    Full Model (Only G; $K_H$ = 12; $K_S$ = 8) & \textbf{0.625 $\pm$ 0.08} & 0.586 $\pm$ 0.08 & \textbf{0.577 $\pm$ 0.10} &  \textbf{0.591 $\pm$ 0.03} & \textbf{0.616 $\pm$ 0.07} \\  
    \bottomrule
  \end{tabular}
\end{table*}

\section{Ablation Studies}
\label{sec:further_ablation}

In this section, we detail the specific ablation settings for PathoSpatial architecture to isolate the contribution of each core component. We also extended the component ablation study to the single-modality baselines to show contribution of each component in a per-modality manner.

\subsection{Ablation Settings}

To systematically evaluate our design choices, we defined three variant configurations. Each variant removes exactly one component while keeping the remainder of the pipeline and hyperparameters (e.g., latent dimension $D=512$) constant.

\textbf{Spatial Dependency Modeling}: This setting evaluates the contribution of the TransMIL-inspired spatial encoder. We hypothesize that modeling spatial context is crucial to yield a contextualized feature set and enhance the discriminative power of the subsequent prototype learning. In this variant, we disable the PPEG module, the Transformer layers, and the residual connection described in Sec. \ref{sec:spatial_dependencies_supp}. The modality-specific embeddings are simply projected to the common dimension $D$ and used directly:
\begin{equation}
\tilde{\mathbf{H}} = \text{LayerNorm}(\mathbf{H})
\end{equation}
where $\mathbf{H}$ is the output of the initial linear projection. This variant treats the slide as an unstructured bag of instances, ignoring relative spatial positions and long-range correlations.

\textbf{Prototype Learning}: This setting evaluates the efficacy of the prototype bottleneck, which summarizes the slide into $K$ compact descriptors. We hypothesize that prototypes reduce noise and computational complexity compared to using all instances. In this variant, we remove the global prototype bank $\mathbf{P}$ and the sparse attention selection mechanism entirely. Instead of aggregating instances into prototypes, we pass \emph{all} spatially encoded instances directly to the fusion module. The candidate set for fusion becomes the stack of all instance features from both modalities:
\begin{equation}
\mathcal{H}_{\text{abl\_proto}} = [\tilde{\mathbf{H}}^{(H)} \mathbin{;} \tilde{\mathbf{H}}^{(S)}] \in \mathbb{R}^{(N_H + N_S) \times D}
\end{equation}
Consequently, the final aggregation is performed over $N_H + N_S$ tokens rather than $K_H + K_S$ prototypes. This mimics standard Attention-based MIL approaches where every patch is a potential contributor to the final slide-level representation.

\textbf{Cross-Modal Gated Attention Aggregation}: This setting evaluates the necessity of the input-dependent Gated Attention network in the fusion stage. We hypothesize that different patients require different emphasis on specific prototypical features, and we seek to learn such proportions via gated attention mechanism. In this variant, we replace the learnable attention network (described in \ref{sec:crossmodal_fusion_supp}) with a static Mean Pooling operation. The fusion weights are fixed to a uniform distribution $\alpha_r = \frac{1}{R}$, where $R = K_H + K_S$. The fused representation becomes:
\begin{equation}
\mathbf{h}_{\text{fused}} = \frac{1}{R} \sum_{r=1}^{R} \mathbf{h}_r
\end{equation}
This forces the model to treat every detected prototype as equally important for the survival prediction, disregarding the specific context or confidence of the detection.

\textbf{Single-Modality Baselines}: To quantify the independent prognostic value of each data source and isolate the performance gain attributable to the cross-modal synergy, we trained the framework using only one modality at a time while keeping the intra-modal architecture identical. \textbf{Histology Only}: The spatial transcriptomics branch is completely disabled. The candidate prototype pool for the fusion stage is restricted to the histological prototypes: $\mathcal{H} = \tilde{\mathbf{P}}^{(H)}$. The model learns to predict survival solely based on morphological patterns. \textbf{Spatial Transcriptomics Only}: The histology branch is disabled. The candidate pool is restricted to the molecular prototypes: $\mathcal{H} = \tilde{\mathbf{P}}^{(S)}$. The model predicts risk utilizing only gene expression signatures and their spatial distribution.

\textbf{Per-Modality Ablations}: To provide a granular view of architectural benefits, we extended the component ablation study to the single-modality baselines. Specifically, we evaluated the removal of the \textit{spatial dependency}, \textit{prototype learning}, and the \textit{cross-modal fusion} within the isolated histology (H-only) and spatial transcriptomics (G-Only) frameworks. This systematic evaluation allows us to quantify the contribution of each component in a per-modality manner. The results are shown in table \ref{tab:ablation_study_only_h} and table \ref{tab:ablation_study_only_g}, respectively.


\subsection{Ablation Results}
We conduct a comprehensive ablation study to validate the contribution of individual architectural components and data modalities. Results are summarized in Table \ref{tab:ablation_study}.

\textbf{Architectural Components Contributions}: The full model achieves the highest performance on RFS, IBFS, and IDFS, demonstrating the efficacy of the complete PathoSpatial framework. Notably, removing the \textit{prototype learning} module consistently degrades performance across all metrics, confirming that summarizing recurring biological patterns is essential for denoising high-dimensional WSI and ST data.

Regarding the fusion mechanism, the learnable \textit{cross-modal attention} is critical for maximizing performance on IBFS and IDFS. However, we observe that for DRFS and OS, replacing attention with a simpler aggregation yields higher scores. We attribute this to the trade-off between model expressivity and generalization, while attention mechanism successfully captures complex, fine-grained cross-modal interactions, a simpler fusion strategy may offer better robustness for endpoints where the signal-to-noise ratio is lower.

Removing the \textit{spatial dependency modeling} component leads to a performance drop across most endpoints. This indicates that simply using pre-trained encodings is insufficient; the model must actively 'blend' global contextual information into local representations to accurately predict complex survival outcomes. However, we observe that OS performance slightly improves without this blending, reinforcing the pattern that overall Survival in this cohort may benefit from simpler feature representations that are less prone to overfitting.

We performed independent ablations within the unimodal histology (Table \ref{tab:ablation_study_only_h}) and spatial transcriptomics (Table \ref{tab:ablation_study_only_g}) baselines. Consistent with the main multimodal results, removing the prototype learning module leads to a degradation in performance across nearly all metrics for both modalities. This confirms that the prototype-based summarization is a fundamental for both morphological and genomic representation learning, independent of fusion. The behavior of the \textit{spatial dependency Modeling} and \textit{cross-modal aggregation} modules also mirror the main ablation study. While these components are crucial for tasks requiring fine-grained discrimination, we observe the same complexity penalty pattern for Overall Survival (OS). Specifically, in the H-only model, removing the feature blend actually improves OS ($0.601 \to 0.610$). This reinforces our hypothesis that for global endpoints like OS, simpler representations may generalize better than complex, spatially-sensitive ones, regardless of whether the model is unimodal or multimodal.

\textbf{Modality Contributions}: Integrating both modalities consistently outperforms single-modality baselines, evidencing the necessity of complementary information. We observe distinct modality dominance depending on the clinical endpoint; for example, removing the Histology modality causes a sharp decline in RFS, suggesting that morphological patterns are the primary driver for recurrence prediction, whereas ST contributes more evenly across endpoints.

\begin{table*}[!t] 
  \caption{Sensitivity Analysis for top-k choice values. The best performance is highlighted in \textbf{bold}.}
  \label{tab:sensitivity_topk}
  \centering
  \begin{tabular}{l|l|ccccc} 
    \toprule
    Strategy & Parameter & DRFS & RFS & IBFS & IDFS & OS \\
    \midrule
    \multirow{4}{*}{Hardcode} 
     & $k_{im} = 40 $  & 0.615 $\pm$ 0.06 & 0.624 $\pm$ 0.06 & 0.584 $\pm$ 0.16 & 0.620 $\pm$ 0.11 & 0.620 $\pm$ 0.04 \\
     & $k_{im} = 60 $ & \textbf{0.637 $\pm$ 0.05} & \textbf{0.645 $\pm$ 0.06} & \textbf{0.610 $\pm$ 0.15} & \textbf{0.647 $\pm$ 0.09} & 0.621 $\pm$ 0.06 \\
     & $k_{im} = 80 $  & 0.635 $\pm$ 0.06 & 0.634 $\pm$ 0.06 & 0.593 $\pm$ 0.16 & 0.647 $\pm$ 0.10 & 0.636 $\pm$ 0.04 \\
     & $k_{im} = 100 $  & 0.634 $\pm$ 0.05 & 0.624 $\pm$ 0.06 & 0.594 $\pm$ 0.16 & 0.645 $\pm$ 0.09 & \textbf{0.641 $\pm$ 0.03} \\
    \midrule
    \multirow{4}{*}{Proportion} 
     & \( \alpha \) = 0.4 & 0.622 $\pm$ 0.05 & 0.635 $\pm$ 0.05 & 0.601 $\pm$ 0.15 & 0.638 $\pm$ 0.10 & 0.617 $\pm$ 0.04 \\
     &  \( \alpha \) = 0.6 & 0.617 $\pm$ 0.05 & 0.637 $\pm$ 0.06 & 0.594 $\pm$ 0.15 & 0.629 $\pm$ 0.09 & 0.613 $\pm$ 0.04 \\
     &  \( \alpha \) = 0.8 & 0.617 $\pm$ 0.05 & 0.632 $\pm$ 0.05 & 0.594 $\pm$ 0.15 & 0.626 $\pm$ 0.09 & 0.613 $\pm$ 0.03 \\
     &  \( \alpha \) = 1.0 & 0.617 $\pm$ 0.05 & 0.637 $\pm$ 0.06 & 0.592 $\pm$ 0.15 & 0.624 $\pm$ 0.09 & 0.602 $\pm$ 0.03 \\
    \midrule
    Trainable &  Init \( \alpha \) = 0.4 & 0.625 $\pm$ 0.08 & 0.586 $\pm$ 0.08 & 0.577 $\pm$ 0.10 & 0.591 $\pm$ 0.03 & 0.616 $\pm$ 0.07 \\
    \bottomrule
  \end{tabular}
\end{table*}

\begin{table*}[!t] 
  \caption{Sensitivity Analysis for number of prototypes. The best performance is highlighted in \textbf{bold}.}
  \label{tab:sensitivity_num_proto}
  \centering
  \begin{tabular}{l|ccccc} 
    \toprule
    Number of prototypes & DRFS & RFS & IBFS & IDFS & OS \\
    \midrule
    $K_H$ = 4; $K_S$ = 6 & 0.625 $\pm$ 0.05 & 0.636 $\pm$ 0.06 & 0.589 $\pm$ 0.15 & 0.620 $\pm$ 0.10 & 0.617 $\pm$ 0.03 \\
    $K_H$ = 12; $K_S$ = 8 & \textbf{0.637 $\pm$ 0.05} & 0.645 $\pm$ 0.06 & 0.610 $\pm$ 0.15 & 0.647 $\pm$ 0.09 & 0.621 $\pm$ 0.06 \\
    $K_H$ = 16; $K_S$ = 24 & 0.628 $\pm$ 0.05 & 0.640 $\pm$ 0.06 & \textbf{0.616 $\pm$ 0.15} & 0.658 $\pm$ 0.09 & \textbf{0.629 $\pm$ 0.04} \\
    $K_H$ = 24; $K_S$ = 32  & 0.631 $\pm$ 0.05 & \textbf{0.646 $\pm$ 0.07} & 0.615 $\pm$ 0.15 & \textbf{0.662 $\pm$ 0.08} & 0.627 $\pm$ 0.04 \\
    \bottomrule
  \end{tabular}
\end{table*}

\begin{table*}[!t] 
  \caption{Sensitivity Analysis for number of gradient accumulation. The best performance is highlighted in \textbf{bold}.}
  \label{tab:sensitivity_num_gc}
  \centering
  \begin{tabular}{l|ccccc} 
    \toprule
    GC & DRFS & RFS & IBFS & IDFS & OS \\
    \midrule
    8 & 0.603 $\pm$ 0.08 & 0.601 $\pm$ 0.06 & 0.603 $\pm$ 0.12 & 0.604 $\pm$ 0.07 & 0.614 $\pm$ 0.07 \\
    16 & 0.605 $\pm$ 0.08 & 0.623 $\pm$ 0.07 & 0.588 $\pm$ 0.15 & 0.633 $\pm$ 0.09 & 0.619 $\pm$ 0.06 \\
    32 & \textbf{0.637 $\pm$ 0.05} & \textbf{0.645 $\pm$ 0.06} & \textbf{0.610 $\pm$ 0.15} & 0.647 $\pm$ 0.09 & 0.621 $\pm$ 0.06 \\
    64 & 0.636 $\pm$ 0.07 & 0.641 $\pm$ 0.07 & 0.603 $\pm$ 0.16 & \textbf{0.652 $\pm$ 0.09} & \textbf{0.635 $\pm$ 0.04} \\
    \bottomrule
  \end{tabular}
\end{table*}

\section{Sensitivity Analysis}
\label{sec:sensitivity_analysis}

In this sensitivity analysis, we try different values for \textit{topk value} for the selection top patches in each bag, \textit{number of prototypes} in each modality and the \textit{number of gradient accumulation} during training.

\noindent \textbf{Top-k selection strategy}: This hyperparameter controls the sparsity of the prototype update and the selectivity of the feature set used for bag prediction. We investigate three strategies: 1) \textit{Hardcoded}, using fixed instance counts $k_{im} \in \{40, 60, 80, 100\}$; 2) \textit{Proportion-based}, where $ k_{im} = \alpha \times (N_i/K_m) $ scales with bag size, testing $\alpha \in \{0.4, 0.6, 0.8, 1.0\}$; and 3) \textit{Trainable}, where $\alpha$ is learned during training. Results in Table \ref{tab:sensitivity_topk} indicate that hardcoded strategies consistently outperform proportion-based and trainable approaches. Specifically, $k_{im}=60$ achieves the best overall performance across most metrics. This suggests that, in this cohort, the number of prognosis-relevant tissue patches or spots is relatively constant and sparse, rather than scaling linearly with the tissue size.

\noindent \textbf{Number of Prototypes}: We vary the size of the prototype bank within each modality to assess the risk of under-clustering (missing patterns) or over-clustering (redundant patterns). As shown in Table \ref{tab:sensitivity_num_proto}, our model demonstrates robustness to the number of prototypes. While larger dictionaries ($K_H=24, K_S=32$) yield marginal gains in RFS and IDFS, the standard setting ($K_H=12, K_S=8$) provides a strong balance of performance and computational efficiency without significant degradation.

\noindent \textbf{Gradient Accumulation}: We analyze the impact of the effective batch size by varying gradient accumulation (GC) steps. Results in Table \ref{tab:sensitivity_num_gc} show that performance is sensitive to the effective batch size. Lower accumulation steps (8 and 16) result in a noticeable performance drop, likely due to noisy gradient estimates from the high variance in MIL bags. Performance stabilizes and peaks at 32 accumulation steps, indicating that a sufficient batch size is critical for learning stable prototype representations.

\section{Model Interpretability}
\label{sec:model_interpretation}

\subsection{Sample-level Interpretability}

\begin{figure*}[t]
  \centering
  \includegraphics[width=1.0\textwidth]{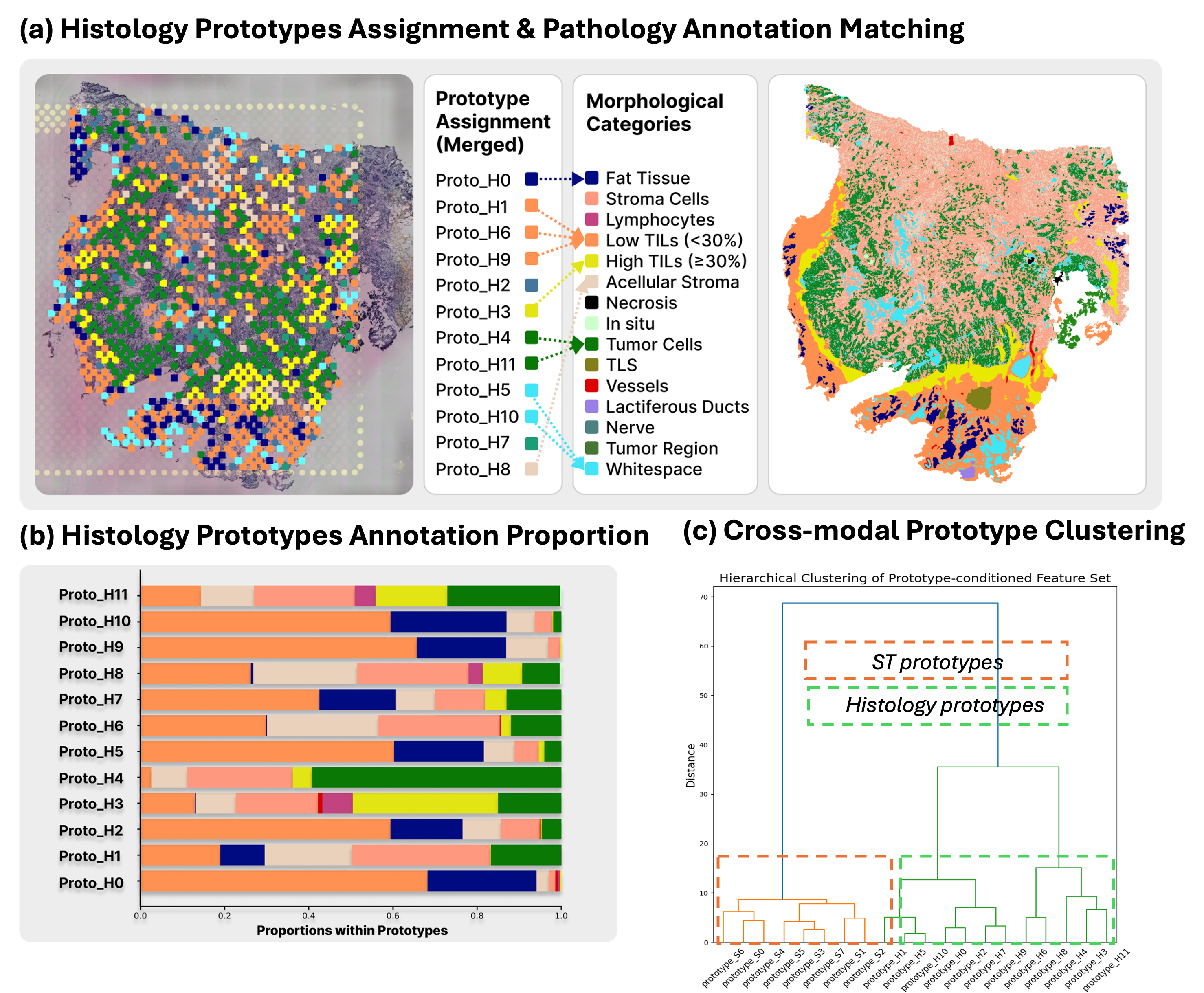}
  \caption{
    \textbf{Extended Model Interpretability}. 
    \textbf{(a)} Visualization of histology prototype assignments alongside ground-truth pathology annotations, demonstrating spatial correspondence between learned concepts and tissue structures. 
    \textbf{(b)} Quantitative analysis of tissue composition for each histology prototype, calculated as the proportion of pathology annotation categories within the top-100 attending patches. This confirms that prototypes specialize in distinct morphological patterns. 
    \textbf{(c)} Hierarchical clustering of prototypes from both modalities. The distinct separation between Histology and ST clusters illustrates the non-redundant, complementary nature of the learned multimodal representations.
    }
  \label{fig:figS1_interpretation}
\end{figure*}

Following the representative case (ID: \textit{TNBC82-CN41-E1}) introduced in the main text, we provide more information about the learned prototypes. This section details the morphological semantics of histology prototypes, the biological pathway enrichment of spatial transcriptomics (ST) prototypes, and the latent structural relationships between the two. Visualizations are provided in Fig. \ref{fig:figS1_interpretation}.

\subsubsection{Histology Prototypes}
To validate the semantic consistency of the learned histological prototypes, we quantified the distribution of pixel-level annotations within the top 100 patches assigned to each prototype. We show the composition of tissue structures (e.g., tumor core, stroma, necrosis, infiltrating lymphocytes) for each prototype-conditioned feature set. For example, \textit{prototype\_H4} and \textit{prototype\_H11} are high in tumor cells (dark green), \textit{prototype\_H3} is the prototype with highest high TILs enrichment (yellow), etc. This breakdown characterizes the distinct morphological patterns captured by the model, confirming that each histology expert specializes in specific, biologically meaningful tissue microenvironments.

\subsubsection{ST Prototypes}
For the Spatial Transcriptomics modality, we characterize the biological function of each prototype by analyzing its associated gene signature. We first identify the molecular drivers of each prototype by performing differential expression analysis (Wilcoxon rank-sum test) between instances attended by the prototype versus the background. Genes are considered statistically significant if they exhibit a Benjamini-Hochberg (BH) adjusted $p$-value $< 0.05$ and a $\log_2$ fold-change $>2$. We then perform Over-Representation Analysis (ORA) to map these statistically significant gene sets to biological pathways using the Gene Ontology (GO) Biological Process (BP) database.

The significance of the enrichment is calculated using a Fisher's Exact Test. Let $N$ be the total number of genes in the background gene (all expressed genes), and $K$ be the number of genes annotated to a specific GO term. If an ST prototype contains a set of $n$ significant genes, of which $k$ overlap with the specific GO term, the probability of observing at least $k$ overlapping genes by random chance is given by the hypergeometric distribution:

\begin{equation}
p = \sum_{i=k}^{\min(n, K)} \frac{\binom{K}{i} \binom{N-K}{n-i}}{\binom{N}{n}}
\end{equation}

We report pathways achieving statistical significance (BH-adjusted $p$-value $< 0.05$), thereby linking the latent ST prototypes to interpretable cellular functions such as immune regulation, cell proliferation, and metabolic activity. Notably, our model captured multiple pathways with established prognostic relevance in TNBC. Supported by prior literature, these include oncogenic drivers such as the MYC \cite{bild2006mycpath} and E2F signaling pathways \cite{bild2006mycpath}, metabolic signatures like Oxidative Phosphorylation (OXPHOS) \cite{elbotty2023oxphos}, as well as structural and immune-related processes including Keratinization \cite{rakha2025keratin} and Allograft Rejection \cite{oshi2022allograft}.

\subsubsection{Inter-modal Prototype Relationships}
Finally, we investigate the latent structure and relationships among the learned prototypes from both modalities. By visualizing the prototype embeddings in a shared projection space, we observe a general separation between histology and ST prototypes. This separation suggests that the two modalities encode distinct, non-redundant information, validating our hypothesis that they provide complementary prognostic signals.

Notably, we observe an exception with \textit{prototype\_S2}, which clusters closer to the histology prototypes than other ST experts. This proximity implies a strong correlation between the molecular program captured by \textit{S2} and the morphological features encoded by the histology encoder, potentially highlighting a specific gene-morphology interplay (e.g., expression signatures heavily driven by morphological changes).


\subsection{Cohort-level Interpretability}

We demonstrate model interpretability at the cohort level by evaluating prototype consistency and modality-oriented patient stratification among the whole cohort.

\subsubsection{Prototype Consistency}
In the pathology modality, the mean adjusted Kappa score across all matched instances achieved $0.68 \pm 0.05$ across five endpoints, especially, tumor (0.85), TIL (0.83), and necrosis (0.78). In the ST modality, prototypes consistently mapped to key biological categories with high prevalence ($>60\%$), such as immune activation ($p < 10^{-12}$), ECM remodeling ($p < 10^{-10}$), and genome integrity ($p < 10^{-9}$). These signals are highly prognostic in breast cancer, underscoring the effective representation learning of our within-modality experts. Note that, the unmatched prototypes may represent novel signals that lie beyond current interpretation, such as fine-grained pathological semantics, topological features, or biological pathways not captured in standard gene set enrichment dictionaries. These signals hold the potential for new hypothesis generation and warrant further investigation.

\subsubsection{Modality-Oriented Patient Stratification} 
\begin{figure}[t]
  \centering
  \includegraphics[width=\columnwidth]{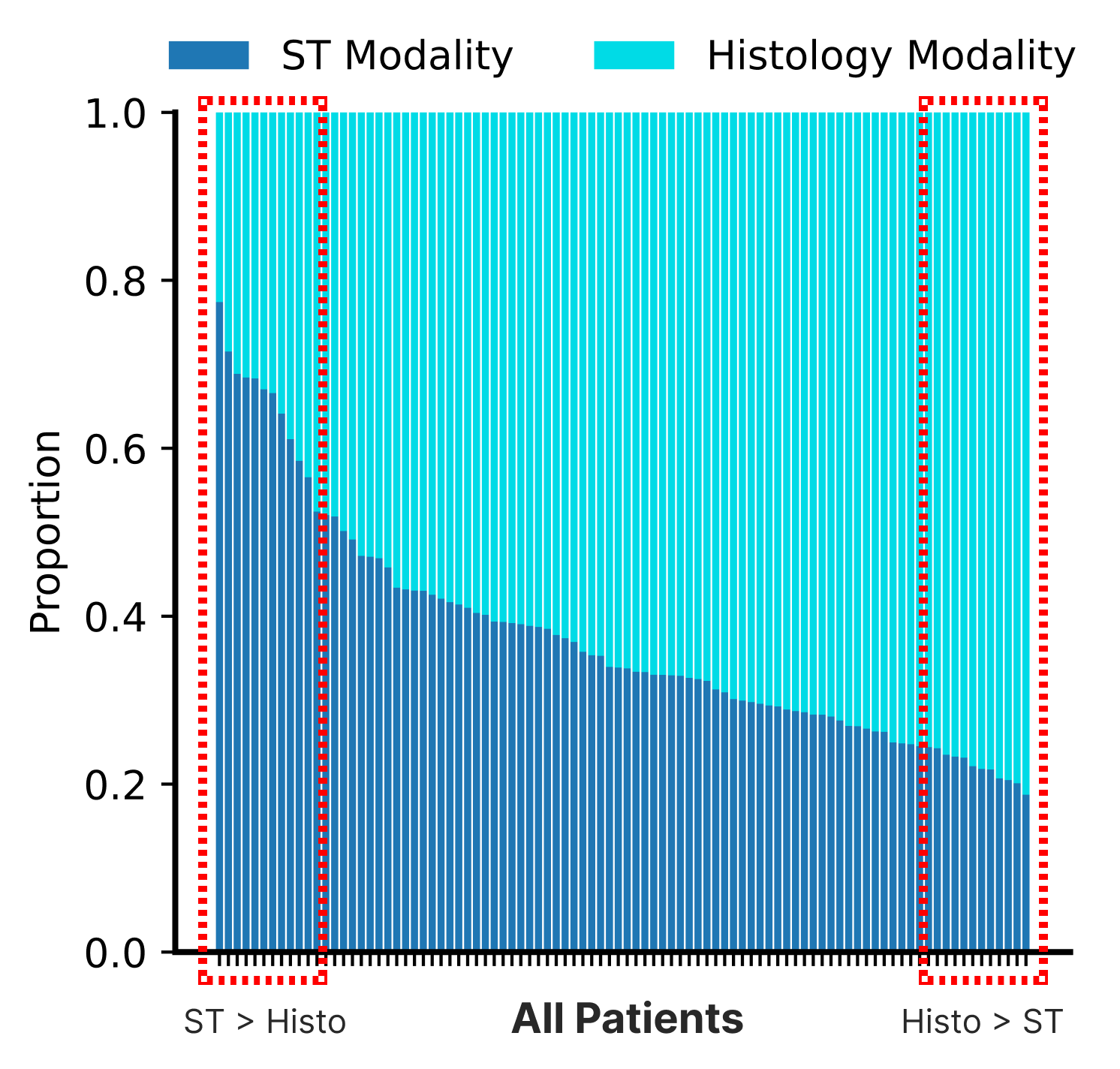}
  \caption{
    \textbf{Modality-Oriented Patient Stratification}.
  }
  \label{fig:figS2_modality_decomp}
\end{figure}

Our model enables a modality-oriented decomposition of risk prediction, see Fig \ref{fig:figS2_modality_decomp}. By aggregating attention scores from each modality, we can decompose patient risk predictions into specific profiles, for instance, identifying cases where the ST modality contribution outweighs H\&E, or vice versa. Such \textit{histo- vs. ST-dominant} patient profiles highlight a non-linear, sample-dependent relationship between modalities and demonstrate the advantage of our adaptive fusion expert. This decomposition may suggest novel stratification with the potential to refine personalized clinical treatment and justifies further exploration in future studies.

\end{document}